\documentclass[twocolumn,floatfix,showpacs,preprintnumbers,amsmath,amssymb]{revtex4-1}

\usepackage{amsmath}
\usepackage{amssymb}
\usepackage{mathrsfs}
\usepackage{bm}
\usepackage{graphicx}
\usepackage{subfig}
\usepackage{algcompatible}
\usepackage{newfloat}
\usepackage[usenames,dvipsnames]{xcolor}
\usepackage{hyperref}
\usepackage[capitalise]{cleveref}

\DeclareFloatingEnvironment[
    fileext=loa,
    listname=List of Algorithms,
    name=ALGORITHM,
    placement=htbp,
]{algorithm}

\newcommand{\fhat}{\hat{f}}

\newcommand{\supp}{\text{supp}}

\newcommand{\Dream}{\emph{Dream} }
\newcommand{\Freedom}{\emph{Freedom} }
\newcommand{\norm}[1]{\left\Vert#1\right\rVert}

\newcommand{\R}{\mathbb{R}}
\newcommand{\N}{\mathbb{N}}

\renewcommand{\S}{\mathbb{S}}

\begin{document}

\title{Modeling a Recurrent, Hidden Dynamical System Using Energy Minimization and Kernel Density Estimates}

\date{(preprint - \today)}

\author{Trevor K. Karn}
\author{Steven Petrone}
\author{Christopher Griffin}
\affiliation{Applied Research Laboratory,
The Pennsylvania State University, University Park, PA 16802, USA}


\begin{abstract}

In this paper we develop a kernel density estimation (KDE) approach to modeling and forecasting recurrent trajectories on a suitable manifold. For the purposes of this paper, a trajectory is a sequence of coordinates in a phase space defined by an underlying hidden dynamical system. Our work is inspired by earlier work on the use of KDE to detect shipping anomalies using high-density, high-quality automated information system (AIS) data as well as our own earlier work in trajectory modeling. We focus specifically on the sparse, noisy trajectory reconstruction problem in which the data are (i) sparsely sampled and (ii) subject to an imperfect observer that introduces noise. Under certain regularity assumptions, we show that the constructed estimator minimizes a specific energy function defined over the trajectory as the number of samples obtained grows. 
\end{abstract}

\maketitle

\section{Introduction}
In this paper we propose an algorithm for modeling and forecasting a sparse, noisy, recurrent trajectory that lies entirely on a smooth Riemannian manifold embedded in an arbitrary dimensional Euclidean space. By \textit{sparse}, we mean the signal may be subject to long gaps in observation; by \textit{noisy} we mean the signal is sampled by an (unknown) imperfect observer. We will define \textit{recurrent} precisely in the context of the underlying mathematical model, however in general we mean the trajectory visits a neighborhood (or collection of neighborhoods) infinitely often. Examples of these trajectories include vehicle (ship, plane and car) tracks, migration data (e.g., in birds, whales and sharks), and some economics data subject to seasonality (e.g., detrended annual sales).

Our approach uses a combination of kernel density estimation (KDE) and energy minimizing inference for sparse trajectory reconstruction prior to model learning. Our goal in using a KDE is to construct distribution estimators, rather than pointwise estimators with confidence intervals. That is, rather than using a traditional pointwise time series forecasting method, our objective is to generate a sequence of probability distributions that can be used to generate an optimized pointwise estimator on demand. The methods proposed in this paper will generalize to smooth Riemannian manifolds in arbitrary dimensions, however we will focus specifically on examples from compact subsets of $\mathbb{R}^2$ and the $2$-sphere $\S^2$ as a representation of the Earth. By way of comparison with similar work \cite{BGH15}, we will show that the proposed approach works well on the attracting set of Lorenz63 \cite{L63}, thus illustrating the proposed approach can generalize outside the assumed smooth Riemannian manifold assumption.

\subsection{Related Work}
Our work extends and is related to the basic statistical problem of time series modeling. Linear and non-linear time series modeling is a well established field of statistics \cite{BJ08} and statistical process control \cite{DelC02}.   Basic linear regression \cite{HT06} (Chapter 8) and non-linear regression \cite{SW03} (Chapter 2) attempt to model observations $\{\mathbf{x}_{t_i}\}_{i=1}^N$ as functions of a variable $t \in \mathbb{R}$. In one dimension, Autoregressive Integrated Moving Average Models (ARIMA) extend these notions by allowing the model $\mathbf{x}_t$ to vary as a function of past values and past shocks \cite{BJ08} (Chapters 3 and 7). Seasonal ARIMA models extend this notion by adding seasonal periodicity \cite{GD01} (Chapter 5). Fractional ARIMA \cite{Hos81} models add short and long range dependence, not expressible with classical ARIMA techniques. In particular, these non-linear models are better able to express persistence and anti-persistence. Finally (Generalized) Autoregressive Heteroskedastic Models (ARCH/GARCH) add heteroskedastic behavior to the error components of the time series, allowing globally stationary and locally non-stationary error terms to be analyzed \cite{Eng01}. Many of these techniques can be extended to vector valued functions (of the type we consider). In particular, Vector Autoregressive models (VAR and VARIMA) \cite{HH08} can be used to model time series of vector valued functions. The most general models are the stochastic differential and difference equations that use Weiner and L\'{e}vy processes to model stochasticity \cite{Ok07} (Chapter 1). Kernel based approaches for forecasting stochastic dynamical systems modeled by (hidden) stochastic differential equations are considered extensively by Giannakis et al. \cite{BGH15,G19}. In particular, in \cite{BGH15,G19} the authors use a diffusion forecasting approach. The shift map of the stochastic process is expressed  in a smooth basis of eigenfunctions. This is used to estimate the semigroup solution of the unknown stochastic differential equation without specific parameter estimation.

Grid-based methods that approximate the trajectory can be employed when the space of the time series is continuous but can be partitioned into a collection of discrete grid points and the trajectory modeled as a time series of these grid points. The work in \cite{MF92, Rab89, SB90} describes methods of using hidden Markov models (and in the case of \cite{MF92}, dynamic programming) to identify optimal estimators for the behavior of trajectories passing through the discretized state space. \cite{GBS11} uses a multi-resolution grid model and a continuous time model to construct a hybrid track estimator that attempts to retain the simplicity of a grid-based model without sacrificing the accuracy of a continuous model. We note that many (but not all) of the approaches discussed are designed to generate pointwise forecasts with confidence regions, while our objective in using a KDE-based approach is to generate a sequence of probability distributions, which can be used to generate a pointwise forecast.

Forecasting dynamical systems, especially non-linear dynamical systems, is a well known problem in physics with applications to noise reduction and experimental data smoothing. Molecular trajectory modeling is considered in \cite{NSVN16,NKPM14} using a variational approach with user supplied basis functions. This approach is in contrast to the the standard Markov process approaches, which are more reminiscent of grid-based methods, already discussed. \cite{KY88,KS93,S93} consider noise reduction in dynamical systems with \cite{S93} providing a fitting approach that is qualitatively similar to the work presented in this paper. Anomaly detection is considered in \cite{CTGP04} with stated goals similar to those in \cite{Pallotta_2013}, but applied to one-dimensional chaotic signals. Forecasting and non-linear modeling is considered in \cite{MS99,V01,DY06,SOSO13,BGH15,G19}. In addition to this work, \cite{LC10} applies stochastic hidden Markov models to fuzzy time series forecasting. Fuzzy time series forecasting is also considered in \cite{HYH07}. \cite{HRXQ18} considers the problem of non-uniform state space reconstruction of chaotic time series. Chaotic time series forecasting is also considered in \cite{SCZC13}, which uses an ant colony optimization algorithm to optimally embed a time series in an appropriate space. Joint continuous and discrete forecasting is considered in  \cite{RRZ13}, while outlier detection of time series is considered in \cite{RA14}. More recent work has applied multi-layer perceptron neural networks to time series forecasting \cite{RIMY16}. 

Using a KDE for the purpose of modeling and forecasting recurrent trajectories has been studied by other authors in more restricted contexts. Pallotta, Vespe, and Bryan \cite{Pallotta_2013} use a kernel density estimation technique to model shipping routes using Automatic Identification System (AIS) data. They use the resulting distributions to identify anomalous behavior in ship routes. As they note, AIS data is exceptionally dense, and can be used in real-time to track ships.  

Additionally, it is well known that a kernel density estimate can be used as a convolutional filter on noisy data. This was done in \cite{streaming_kde} in order to visualize streaming data from an aircraft. This is a trajectory in $\R^3$, although \cite{streaming_kde} only considers the projection onto $\R^2$. In both \cite{Pallotta_2013} and \cite{streaming_kde}, the data are highly dense with minimal noise. This is not realistic in antagonistic situations or in cases where the trajectory cannot be observed with high fidelity. This occurs naturally when biologists observe animals in their natural habitat (e.g., see \cite{OC18}). This paper considers situations in which the sampled trajectory is neither dense nor exhibits high signal-to-noise ratio (SNR). We contrast this to the work in \cite{G19,BGH15}, where the data are assumed to be more dense.

\subsection{Paper Organization}
The remainder of this paper is organized as follows: In \cref{sec:Notation} we introduce notation and the underlying mathematical model to be used throughout the rest of the paper. In \cref{sec:Alg} we discuss our proposed modeling and forecasting algorithms. Theoretical results on the algorithms are provided in \cref{sec:theo_results}. We present empirical results using synthetic and real-world data sets in \cref{sec:Empirical}. Finally conclusions and future directions are presented in \cref{sec:Concl}.

\section{Notation and Preliminaries}\label{sec:Notation}

\subsection{Notation and Assumptions}
Let $\mathbb{R}$ denote the real numbers, and ${\mathbb{R}^+ = \{x \in \mathbb{R}: x \geq 0\}}$.
Let $\langle{M,\mathbb{R}^+,\varphi}\rangle$ be a (hidden) dynamical system describing the motion over time ($\mathbb{R}^+$) of an (autonomous) particle on a $d$-dimensional, smooth Riemannian manifold $M$. The manifold $M$ may be embedded in Euclidean space of dimension $m\leq 2d$, and such an embedding is guaranteed to exist by Whitney's Strong Embedding Theorem. Throughout this paper, boldface symbols will indicate positions on the manifold in an appropriate coordinate system; e.g., $\mathbf{x} \in \mathbb{R}^d$ and $\mathbf{x} = \langle{x_1,\dots,x_d}\rangle$. 
We will often (without explicitly stating) identify the set of points in $M$ with their image in $\mathbb{R}^d$ under an appropriate chart.

If $M$ is known, then our approach may be taken using $M$ itself, e.g. using the KDE theory developed in \cite{brigant_puechmorel_manifolds}. In the case $M$ is unknown, but the image of the embedding $M \hookrightarrow \mathbb{R}^m$ is known, our approach may be used taking $\mathbb{R}^m$ to be the manifold of interest (even though the data may be drawn from a different underlying manifold). Determining $M$ given data in $\mathbb{R}^m$ is the fundamental problem in topological data analysis \cite{G07,C09} and will not be explored further here.

Since our manifold is Riemannian, it may be endowed with an appropriate metric. For example, when $M \equiv \mathbb{R}^d$, the standard Euclidean metric is used; when $M \equiv \mathbb{S}^2$ (the 2-sphere), the Haversine metric is applicable. We denote distance between two points $\mathbf{x},\mathbf{y}$ in $M$ as $d(\mathbf{x},\mathbf{y})$. 
Again by our assumption of a Riemannian manifold, we have the existence of an inner product (positive-definite metric tensor), denoted $\langle \mathbf{x},\mathbf{y}\rangle$. This should not be confused with a two-dimensional vector as the entries are vectors in bold typeface rather than coordinates in standard typeface. We will use the inner product to quantify the degree to which (e.g., velocity) vectors located at $\mathbf{x},\mathbf{y}$ have similar heading. In Euclidean space, we would choose the usual dot product.
Throughout the paper, we use $\overset{\Delta}{=}$ to denote equality by definition rather than derivation.

The dynamical system we study is hybrid in the following sense: Fix a finite set $\mathscr{O} \subseteq M$. At any time $t$, either:
\begin{enumerate}
\item There are positions $\mathbf{x}_0, \mathbf{x}_f \in \mathscr{O}$ and the function $\varphi$ defines a sub-trajectory $\mathbf{x}_t = \varphi_{\mathbf{x}_0}(t - t_0)$ in $M$ so that
\begin{equation}
\left\{
\begin{aligned}
\varphi_{\mathbf{x}_0}(t-t_0) \overset{\Delta}{=} \arg\min_\varphi  & \;\; \int_{t_0}^{t_f}\mathscr{L}(\varphi,\dot{\varphi},t) \; dt\\
s.t.\;\; & \mathbf{g}_t(\varphi,\dot{\varphi}) \leq \mathbf{0}\\
&\varphi_{\mathbf{x}_0}(t_0) = \mathbf{x}_0 \\
&\varphi_{\mathbf{x}_0}(t_f - t_0) = \mathbf{x}_f
\end{aligned}
\right.
\label{eqn:EnergeyMinDyn}
\end{equation}
where $\mathscr{L}:(\varphi,\dot{\varphi},t) \mapsto r \in \mathbb{R}$ is a hidden energy function (Lagrangian) and $\mathbf{g}_t:(\varphi,\dot{\varphi}) \mapsto \mathbf{b} \in \mathbb{R}^m$ are hidden (possibly time parameterized) constraints.

\item We have $\varphi(t) = \mathbf{x}_0 \in \mathscr{O}$. At some time $t + \tau$, a new $\mathbf{x}_f \in \mathscr{O}$ is chosen (possibly at random). 
\end{enumerate}
We assume the dynamical system is recurrent in the sense that the choice of $\mathscr{O}$ is governed by an ergodic or periodic (hidden) Markov chain with no transient states. Therefore, if $\varphi(t) = \mathbf{x}_0$, there is some $T < \infty$ so that $\varphi(t + T) = \mathbf{x}_0$.

In the problem we discuss, all relevant information about the dynamics, including $\mathscr{O}$, the Lagrangian $\mathscr{L}$ and some (perhaps all) of the constraints $\mathbf{g}_t$ are hidden. The assumption that $\varphi(t)$ is constructed from piecewise optimal paths is used to justify our method of inferring missing information in sparsely sampled data. For simplicity, in the remainder of this paper, we will assume that $\mathbf{g}_t(\varphi,\dot{\varphi})$ are time invariant and denote the constraint functions by  $\mathbf{g}$. 
In the sequel, we assume data are sampled discretely via a sampling function $\eta:M \rightarrow \mathbb{R}^d$ (or $\eta:M \rightarrow \mathbb{R}^m$ as appropriate) and with unbiased noise to produce a sparse noisy signal:
\begin{equation}
  \mathbf{x}_{i} = \eta(\varphi(t_i)) + \bm{\epsilon}_{t_i}
  \label{eqn:Position}
\end{equation}
here the $\bm{\epsilon}_{t_i}$ are unbiased noise vectors of appropriate dimension. In the sequel, we will elide the observation function $\eta$ for the sake of clarity and identify $\varphi(t_i)$ with $\eta \circ \varphi(t_i)$. Whether we are using $\varphi$ to mean a trajectory on $M$ or its image in $\mathbb{R}^d$ or $\mathbb{R}^m$ will be clear from context.

We note, our approach is a parameter-free approximation method and our focus is \textbf{not} on estimating the distribution 
that describes $\bm{\epsilon}_{t_i}$, unlike (e.g.) in the traditional Kalman filter estimation (see \cite{kalman_filter}).

In addition to the recurrence assumption, we assume that $\varphi(t)$ is piecewise smooth, and in particular at $t_0$ an instantaneous velocity can be constructed using initial conditions. In practice velocity is numerically approximated by a difference quotient. Finally, since we assume that $\varphi_{\mathbf{x}_0}(t - t_0)$ obeys a set of externally imposed constraints defined by $\mathbf{g}(\varphi,\dot{\varphi})$ in \cref{eqn:EnergeyMinDyn}, we assume there is a feasible region $\Omega \subseteq M$ defined by $\mathbf{g}(\varphi,\dot{\varphi})$ and for all $t$, $\varphi(t) \in \Omega$. In particular, $M$ may be convex as a set in $\mathbb{R}^m$, but $\Omega$ may not be, making the problem more challenging.

\subsection{Techniques}\label{ssec:Techniques}

We provide a brief overview of Kernel Density Estimates (KDE) in Euclidean space, which are a foundational element of our proposed algorithm. The interested reader may consult \cite{mvde_scott} for a more detailed overview of Euclidean KDE methods, or \cite{brigant_puechmorel_manifolds} for KDE methods on a Riemannian manifold.

The KDE is a non-parametric estimate of the probability distribution of a set of data $\{x_i\}_{i=1}^N$. 
Formally, the univariate KDE $\hat{f}$ is given by
\begin{displaymath} 
\hat{f}(x) = \frac{1}{Nh} \sum \limits_{i=1}^N K \left ( \frac{x-x_i}{h} \right),
\end{displaymath}
with the requirement that the kernel $K$ be a valid probability density function (PDF). Intuitively, one may think of a KDE as a mean of many probability distribution functions with a normalizing constant, $\frac{1}{h}$. The parameter $h > 0$ is referred to as the \textit{bandwidth} and there are many well established rules-of-thumb for choosing $h$ given arbitrary data (see, e.g., \cite{JMS96}). The bandwidth controls the variance of the kernel and acts as a smoothing control on the resulting KDE.

The KDE can be generalized to arbitrary-dimensional Euclidean space. Let $\{\mathbf{x}_i\}_{i=1}^N$ be $N$ data points in $\mathbb{R}^m$. Define $\mathbf{x}_i = \langle{x_{i_1},\dots,x_{i_m}}\rangle$. The KDE generalizes to:
\begin{displaymath}
\hat{f}(\mathbf{x}) = \frac{1}{N h_1 h_2 \cdots h_m } \sum \limits_{i = 1}^N \left ( \prod \limits_{j = 1}^m K \left( \frac{x_j - x_{i_j}}{h_j}\right)\right ),
\end{displaymath} 
where $K$ is a PDF defined on $\mathbb{R}^d$ and $h_j$ is the bandwidth parameter for the coordinate $x_j$. We will write the \emph{bandwidth vector} $\mathbf{h} ={ \langle h_1, h_2, ..., h_d \rangle} \in (\R^+)^d$. The product defined inside the sum is called the \textit{product kernel}

\textbf{Choice of kernel:} One consideration which must be taken into account when using KDE methods is the choice of kernel. 
While the only restriction on $K$ is that it is a valid PDF, there are a few canonical choices of kernel.
The first, often used in image processing applications as the kernel of a convolutional filter on images, is the Gaussian/normal PDF.
Moreover, the Gaussian kernel is used in both \cite{Pallotta_2013} and \cite{streaming_kde}.

Another canonical choice is the Epanechnikov kernel. Defined as 
\[ K(x) = \begin{cases}\frac{3}{4}(1 - x^2)& x \in [-1,1]\\ 0 &\text{otherwise}\end{cases},\]
the Epanechnikov kernel is a compactly supported parabola.  
In \cite{E69}, Epanechnikov shows that the kernel minimizes relative global error in the following sense: Let $f$ be the true distribution and assume that $f$ is analytic. Suppose $f$ is approximated by $\hat{f}$ using Epanechnikov kernel. Then $\hat{f}$ minimizes the functional:
\begin{equation}
\iint_{\mathbb{R}^2} \mathbb{E}[\hat{f}(\mathbf{x}) - f(\mathbf{x})]^2 \;d\mathbf{x},
\label{eqn:MISE}
\end{equation}
where $\mathbb{E}$ is the classical expectation operator. 
The preceding result is derived using the calculus of variations with constraints and is independent both of distribution (subject to the requirement of analyticity) and constraints on the kernel function.

\cref{eqn:MISE} defines the  Mean Integrated Square Error (MISE). Therefore, the 
Epanechnikov kernel minimizes MISE. Scott \cite{mvde_scott} notes ``...the MISE error criterion has two different, though equivalent, interpretations: it is a measure of both the average global error and the accumulated point-wise error." The fact that the Epanechnikov kernel is a MISE minimizer makes it the natural choice of non-Gaussian kernel. 

Moreover, the fact that the Epanechnikov kernel has compact support also implies that the inferred distribution has compact support (as the union of finitely many compact sets), which is sometimes desirable.
In the context of our work we may implicitly constrain the velocity of the forecast by requiring the KDE have a compact support. The Gaussian kernel does not respect such a constraint, as it gives non-zero probability to the whole ambient space, rather than restricting the non-zero probability to a region that satisfies velocity constraints. As we also show in \cref{sec:theo_results}, a kernel with bounded support can also provide meaningful results regarding feasible regions.

\textbf{KDE in the Plane:}
In $\R^2$, which is a sufficient for approximation of the earth's surface over small regions, we may choose the Epanechnikov product kernel, given by the expression:
\begin{multline*}
\hat{f}(x,y) = \\
\frac{9}{16N h_1 h_2} \sum \limits_{i = 1}^N   \left(1-\frac{(x-x_i)^2}{h_1^2}\right)\left(1-\frac{(y-y_i)^2}{h_2^2}\right) . 
\end{multline*}
This is our choice of kernel for the experimental results of Section \ref{sec:SynthForecast}.

\textbf{KDE on the Sphere:}
In \cref{sec:Empirical}, we apply the proposed algorithm to cruise ships traveling on the surface of the earth. While for this paper we consider a small enough region to approximate by the Earth as a plane, on larger scales this may lead to significant distortion. In such a case one might wish to approximate the Earth as a 2-sphere $\S^2$. In this case, one could use use the Kent distribution, the analogue to the Gaussian distribution on a sphere, as the choice of kernel. Let $\lambda$ be the longitude in degrees, and $\phi$ the the latitude in degrees. The general formulation of the Kent distribution in spherical coordinates is
\begin{displaymath}
f( \lambda, \phi) = c(\kappa, \beta)^{-1} \exp \left ( \kappa \cos \lambda + \beta \sin^2\lambda \cos 2\phi \right).
\end{displaymath}
Here $\kappa$ is a parameter representing the \textit{concentration} of the distribution, $\beta$ is an analogue of covariance, which Kent describes as the ``ovalness," and $c(\kappa, \beta)$ is a normalizing constant given by 
\begin{displaymath}
c(\kappa, \beta) = \int \limits_{0}^{\pi} \int \limits_{0}^{2\pi} \exp \left( \kappa \cos x + \beta \sin^2 x \cos 2y \right) \sin x dy dx.
\end{displaymath} 
We make the simplifying assumption that the covariance of $x,y$ is $0$, which is equivalent to setting $\beta = 0$.
Then we have $c(\kappa, 0) = 4 \pi \kappa^{-1} \sinh \kappa$, simplifying the double integral above. A full description of the Kent distribution can be found in \cite{FBdist_kent}. It is also possible to find a KDE on other compact Riemannian manifolds without boundary (see  \cite{pelletier_kde_manifolds,brigant_puechmorel_manifolds}). We do not further discuss this case, as Euclidean space is sufficient for our (and most other) applications.

\subsection{Deriving a Point Estimator and Uncertainty Regions with a KDE}
Given a PDF (potentially a KDE) $f$ on $M$, there are several ways to find a point estimator. The most obvious method is to compute the argument of expectation, $\mathbf{e}\in M$ such that $f(\mathbf{e}) = \mathbb{E}[f]$. However, if $f$ is multimodal on $M$ and there are constraints on the dynamics (see \cref{eqn:EnergeyMinDyn}), then it is possible $\mathbf{e} \not\in \Omega$, where $\Omega \subseteq M$ is the feasible subset of $M$. In this case it is more useful to to compute:
\begin{equation}
\tilde{\mathbf{x}} = \arg \max_{\mathbf{x} \in \Omega} f(\mathbf{x}),
\label{eqn:argmaxx}
\end{equation}
as the point estimator. Depending on the numerical complexity, one can also define the conditional distribution $f_{\mathbf{x}|\Omega}$ and compute $\mathbf{e},\tilde{\mathbf{x}}$ accordingly. We note that \cref{eqn:argmaxx} may be a non-convex optimization problem. In this case, it might be simpler to compute the unconstrained optimization: 
\begin{equation}\tilde{\mathbf{y}} = \arg\max_{\mathbf{y} \in M} f(\mathbf{y}) \label{eqn:pcomp}\end{equation}
and either (i) accept that $\tilde{\mathbf{y}}  \not \in \Omega$ or (ii) project $\tilde{\mathbf{y}} $ onto the boundary of $\Omega$. We discuss the limits of this approach in \cref{sec:theo_results}.

To find uncertainty regions, we compute the highest density region, following the Monte Carlo technique established in \cite{hdr_hyndman}. Intuitively, the \textit{highest density region} (HDR) of a distribution function $f:M \rightarrow \mathbb{R}^+$ is the subset of $M$ corresponding to the preimage of a horizontal ``slice'' of ${\R^+ }$
where the slice includes the maximum of the PDF, and continues extending down until the probability measure of the preimage of the slice meets a predetermined threshold.

More formally, given a PDF $f$ with support $X \subseteq M$ define:
\begin{equation}
R(c) = \{\mathbf{x} \in X: f(\mathbf{x}) \geq c\}.
\label{eqn:RegionComp}
\end{equation} 
The $100\% \times (1-\alpha)$ highest density region is the set $R(c_\alpha)$ where: 
\begin{equation}
c_\alpha = \arg\max_{c} \idotsint_{R(c)} f(\mathbf{x}) \; d\mathbf{x} \geq (1-\alpha).
\end{equation}
It is clear from this definition that $\tilde{\mathbf{y}} \in R(c_\alpha)$ for any $0 \leq \alpha < 1$. If we compute the HDR of the conditional distribution $f_{\mathbf{x}|\Omega}$, then the probability measure of $M \backslash \Omega$ is zero, and so $\tilde{\mathbf{x}} \in R(c_\alpha)$. This is also the case if the unconditional probability given to $M \backslash \Omega$ is sufficiently small. 

Hyndman \cite{hdr_hyndman} proposes a Monte Carlo algorithm that samples the computed PDF (in our case the KDE) and uses the $\alpha$ quantile as an estimator for $c_\alpha$. We use this algorithm in the sequel. 

Lastly, since we consider a sequence of KDE's that advance in time, there is an implicit inclusion of the velocity of the object. Thus, we do not need to actually approximate or compute $\dot{\varphi}_{\mathbf{x}_0}(t - t_0)$ explicitly while generating a forecast. This stands in direct contrast to alternate approaches (e.g., \cite{G19,BGH15,NKPM14,NSVN16}), which view forecasting as finding and solving a system of stochastic differential equations describing the motion of a particle.

\section{Algorithm Description}\label{sec:Alg}
In this section we motivate Algorithm \ref{algo:Main}, which forecasts a finite sequence of triples $\mathscr{F} = \{(\mathbf{p}_i,\hat{f}_i,R_i)\}_{i=N+1}^{N+\overline{q}}$, with $\overline{q} \in \N$, where $\mathbf{p}_i$ estimates the value $\varphi_{\mathbf{x}_0}(t_i-t_0)$ and $\hat{f}_i:M \to \mathbb{R}$ is a distribution used to construct the HDR $R_i$ that acts as an uncertainty region for $\mathbf{p}_i$. The algorithm takes as input the observation set $P=\{\mathbf{x}_i\}_{i=1}^N$, a time-indexed sequence of observed positions. Recall from  \cref{eqn:Position}, $\mathbf{x}_i$ is observed with noise. Note that we do not require $t_i - t_{i-1} = t_{i+1} - t_i$ for the points in $P$. 

The algorithm is broken into four stages:
\begin{enumerate}
\item Identify a collection of points 
$${H = \{\mathbf{x}_{i_j} \in P : j \in J \}}$$
\textit{similar} (defined by the metric and/or inner product on $M$) to the last known state of the particle. The set $J$ is an index set of consecutive integers which respect the time series $P$, that is $i_j < i_{j+1}$ for all $j$. 

\item Extract sub-trajectories of $P$ corresponding to each identified point in $H$ that (i) begin at the identified point and (ii) span an input forecast time. This set of sub-trajectories is denoted $\mathscr{P}$.

\item Densify the observed sub-trajectories in $\mathscr{P}$ to obtain $\overline{\mathscr{P}}$ using a line integral minimization on an estimator $\hat{\mathscr{L}}$ of $\mathscr{L}$. Each densified trajectory in $\overline{\mathscr{P}}$ is composed of points on $M$ that are equally spaced in time.

\item Let:
\begin{displaymath}
\overline{\mathscr{P}} = \left\{ \left\{\overline{\mathbf{x}}_i^{(j)}\right\}_{i=i_j}^{i_j + \overline{q}} : j \in \{1,\dots,|H|\}\right\}
\end{displaymath}  
be the densified trajectories. For each $i\geq N$ (time index) use the set:
\begin{displaymath}
X_i = \left\{\overline{\mathbf{x}}_{i_j+(i-N)}^{(j)} :  j \in \{1,\dots,|H|\}, i_j+i \leq N \right\}
\end{displaymath}
to construct a KDE $\hat{f}_i$. Use the KDE to construct $\mathbf{p}_i$ and an associated HDR representing an uncertainty region. 
\end{enumerate}

\noindent\textbf{Metrics and Tolerances:}
In Section \ref{sec:Notation}, we have already defined the distance $d(\mathbf{x},\mathbf{y})$  and inner product $\langle \mathbf{x}, \mathbf{y}\rangle$ on the manifold $M$. Given two velocity vectors $\mathbf{v}_i$ and $\mathbf{v}_j$, let the angle metric be:
\begin{displaymath}
\delta(\mathbf{v}_i,\mathbf{v}_j) \overset{\Delta}{=} 
	1 - \frac{\langle{\mathbf{v}_i,\mathbf{v}_j}\rangle}
	{\norm{\mathbf{v}_i}\norm{\mathbf{v}_j}}
\end{displaymath}
This is just the standard cosine distance when $M \equiv \mathbb{R}^d$. In the absence of velocity data, we can use a finite difference method so that (e.g.) in $\mathbb{R}^d$, the velocity vector of the last observed point in $P$ is:
\begin{displaymath}
\mathbf{v}_{N} \approx \frac{\mathbf{x}_N - \mathbf{x}_{N-1}}{t_{N} - t_{N-1}}.
\end{displaymath}

More generally, if we are working in $M$, then we may use the language of velocity vectors on smooth manifolds (see e.g. \cite{boumal2010discrete} or Chapter 3 of \cite{lee_smooth}) to approximate $\mathbf{v}_N$. It is in these details that we wish for $M$ to be smooth. The details of this computation obfuscate the presentation of the proposed algorithm, thus we omit them and refer the readers to the provided references. 

As input to Algorithm \ref{algo:Main}, we take two parameters $\epsilon  > 0$, which is a a tolerance on $d(\cdot,\cdot)$ and $\vartheta$, which is a tolerance on $\delta(\cdot,\cdot)$. These function as hyper-parameters in our proposed algorithm.

\noindent\textbf{Forecast Duration:}
Define $P(\mathbf{x}_i,T)$ to be the \textit{forward time restriction of $P$} beginning with $\mathbf{x}_i \in P$  and including all points $\mathbf{x}_k$ so that $t_k - t_i \leq T$. That is:
\begin{displaymath}
P(\mathbf{x}_i,T) = \left\{\mathbf{x}_k \in P : k \geq i \wedge t_k - t_i \leq T \right\} 
\end{displaymath}
In Algorithm \ref{algo:Main}, $T$ is the duration of the forecast and is an input parameter.

\noindent\textbf{Sampling Period:}
For sparse track reconstruction and forecast generation, we require a sampling frequency. The sampling frequency is a value $\Delta t$ so that if $\overline{Q} = \left\{\overline{\mathbf{x}}_i ^{(j)}\right\}_{i=i_j}^{i_j + \overline{q}} \in \overline{\mathscr{P}}$ is a densified trajectory corresponding to some subtrajectory $Q \in \mathscr{P}$, then for all ${i \in \{i_j,\dots,i_j+\overline{q}\}}$: $t_{i+1} - t_i = \Delta t$, where $\overline{\mathbf{x}}_i^{(j)} \in \overline{Q}$. This sampling period gives resolution to intermediate points of prediction but does not affect predictions made at any given point. It is now straightforward to see that $\overline{q} = \lceil T/\Delta t\rceil$.

\noindent\textbf{Track Densification:}
Suppose $\mathbf{x}_i \in H$ and $P(\mathbf{x}_i,T)$ must be \textit{densified}; i.e., there is some pair $\mathbf{x}_j, \mathbf{x}_{j+1} \in P(\mathbf{x}_i,T)$ so that so that $t_{j+1} - t_j > \Delta t$. (Note: if $P(\mathbf{x}_i,T)$ is too dense, it is trivial to downsample it to make it sparser.) If approximations $\hat{\mathscr{L}}$ and $\hat{\mathbf{g}}$ are available, then it is trivial to solve (numerically):
\begin{equation}
\left\{
\begin{aligned}
\min_{\varphi}  & \;\; \int_{t_j}^{t_{j+1}}\hat{\mathscr{L}}(\varphi,\dot{\varphi},t) \; dt\\
s.t.\;\; & \hat{\mathbf{g}}(\varphi,\dot{\varphi}) \leq \mathbf{0}\\
& \varphi(t_j) = \mathbf{x}_j,\;\varphi(t_{j+1}) = \mathbf{x}_{j+1}
\end{aligned}
\right.
\label{eqn:Densify}
\end{equation}
The resulting solution can be used to provide an estimated track of arbitrary density. If $\hat{\mathscr{L}}$ is not already available, then it is straight-forward to define a Gaussian well function:
\begin{equation}
\hat{\mathscr{L}}_G(\mathbf{x}) \overset{\Delta}{=}  \sum_{j=1}^N \left(1-\frac{1}{\sqrt{2\pi\sigma_i^2}}\exp\left(-\frac{d(\mathbf{x},\mathbf{x}_i)^2}{2\sigma_i^2}\right)\right)
\label{eqn:GaussianWells}
\end{equation}
or a least squares function:
\begin{equation}
\hat{\mathscr{L}}_{LS}(\mathbf{x}) \overset{\Delta}{=}  \sum_{j=1}^N \frac{1}{\sigma_i} d(\mathbf{x},\mathbf{x}_i)^2
\label{eqn:LSWells}
\end{equation}
and solve the constrained line integral minimization problem:
\begin{equation}
\left\{
\begin{aligned}
\min_{\varphi}  & \;\; \int_{\varphi}\hat{\mathscr{L}}_{*}(\mathbf{x}) \; d\mathbf{x}\\
s.t.\;\; & \hat{\mathbf{g}}(\varphi,\dot{\varphi}) \leq \mathbf{0}\\
& \varphi(t_j) = \mathbf{x}_j,\;\varphi(t_{j+1}) = \mathbf{x}_{j+1}
\end{aligned}
\right.
\label{eqn:MinLhat}
\end{equation}
Here, $*$ indicates the choice of Lagrangian. Using \cref{eqn:GaussianWells} in \cref{eqn:Densify} causes inferred trajectories to follow historical trends, since the center of the Gaussian well yields the minimal energy, while using \cref{eqn:LSWells} causes the path to minimize the square error. One benefit to  \cref{eqn:LSWells} is it has simpler theoretical properties as we show in the sequel. On the other hand, when using \cref{eqn:GaussianWells}, if $\sigma_i$ is an increasing function of $t_i$, then this is a continuous variant of pheromone routing \cite{SS03,HZCL10}. 

Constraint inference is more difficult. In practical situations (e.g., ship tracks) there are obvious constraints in play, like land avoidance (see \cref{sec:Empirical} for examples). For the remainder of this paper, we assume that the constraint function $\mathbf{g}$ (or at least $\hat{\mathbf{g}}$, and hence the feasible region $\Omega$) is known and consider constraint estimation as future work. Algorithm \ref{algo:Main} shows the pseudo-code for the proposed algorithm. 

\begin{algorithm}
\begin{flushleft}
\textbf{Input:} 
	$P=\{\mathbf{x}_i\}_{i=1}^N$, 
	$\epsilon > 0$, 
	$\vartheta \in [0,1]$, 
	$\Delta t$, 
	$T$,
	$\hat{\mathscr{L}}$, $\mathbf{g}$,
	$\alpha \in [0,1]$\\
\textbf{Output:} 
$\mathscr{F} = \{(\mathbf{p}_i,\hat{f}_i,R_i)\}_{i=N+1}^{N+\overline{q}}$\\
\textbf{Initialize:} 
	$H \leftarrow \emptyset$, 
	$\mathscr{P} \leftarrow \emptyset$,
	$\overline{\mathscr{P}} \leftarrow \emptyset$,
	$\bar{q} \leftarrow \lceil{T/\Delta t}\rceil$
\end{flushleft}

\begin{algorithmic}[1]
\FOR{$\mathbf{x}_i \in P$}
\State \textit{$\triangleright$ Stage 1: Collect start points.}
\IF{$d(\mathbf{x}_i, \mathbf{x}_N) < \epsilon$ {\bf and}  
			$d(\mathbf{x}_{i-1}, \mathbf{x}_N) \geq \epsilon$  {\bf and}
			$\delta(\mathbf{v}_i,\mathbf{v}_N) < \vartheta$ {\bf and}
			$t_N - t_i > T$
			}
\State $H\leftarrow H \cup \{\mathbf{x}_i\}$
\Comment{\textit{Retain index $i$ in $H$.}}
\ENDIF
\ENDFOR

\FOR{$\mathbf{x}_i \in H$}
\State \textit{$\triangleright$ Stage 2: Build sample paths.}
\State $\mathscr{P} \leftarrow \mathscr{P} \cup \{ P(\mathbf{x}_i,T )\}$
\ENDFOR

\FOR{$Q \in \mathscr{P}$}
\State \textit{$\triangleright$ Stage 3: Densify all paths using \cref{eqn:Densify}.}
\State $\overline{Q} \leftarrow \texttt{Densify}(Q)$
\State $\overline{\mathscr{P}} \leftarrow \overline{\mathscr{P}} \cup \{ \overline{Q} \}$
\ENDFOR

\State \textit{$\triangleright$ Note: $\overline{\mathscr{P}} = 
		\{\overline{Q}_1,\dots,\overline{Q}_{|H|}\}$. 
		Also, for each $j \in \{1,\dots,|H|\}$,
		$\overline{Q}_j = 
		\left\{\overline{\mathbf{x}}_i^{(j)}\right\}_{i=i_j}^{i_j+\overline{q}}$.}
		
\FOR{$i \in \{N+1,\dots,N+\bar{q}\}$}
\State \textit{$\triangleright$  Stage 4: Compute $\hat{f}_i$ and $\mathbf{p}_i$.}
\State $X_i \leftarrow \emptyset$
	\FOR{$j \in \{1,\dots,|H|\}$}
		\State $X_i \leftarrow X_i \cup \left\{\mathbf{x}_{i_j+i-N}^{(j)}\right\}$
	\ENDFOR
	\State Compute $\hat{f}_i$ using $X_i$ and a kernel density estimate
	\State Compute $\mathbf{p}_i$ using \cref{eqn:pcomp}
	\State Compute $R_i$ using \cref{eqn:RegionComp}
\ENDFOR
\end{algorithmic}
\caption{Forecasting Algorithm}\label{algo:main}
\label{algo:Main}
\end{algorithm}

\section{Theoretical Results}\label{sec:theo_results}
If the Lagrangian $\mathscr{L}$ is stationary, we can show that the optimal solution 
to the problem given in \cref{eqn:MinLhat} is asymptotically $\varphi(t)$ when $t \in [t_0,t_f]$ and $\varphi(t) \equiv \varphi_{\mathbf{x}_0}(t - t_0)$ as the sampling rate increases. Assume $\varphi_{\mathbf{x}_0}(t_f - t_0) = \mathbf{x}_f \in \mathscr{O}$. Further assume we have $n \in \mathbb{N}$ observations of the continuous path connecting $\mathbf{x}_0$ with $\mathbf{x}_f$ denoted as $\{\mathbf{x}^{(i)}(t)\}_{i = 1}^{n}$ with $t_i$ representing the time at which position is observed. We are considering the asymptotic case when the sampling rate is infinite (i.e., $\mathbf{x}^{(i)}$ can be thought of as a function from $[t_0,t_f] \rightarrow M$), so:
\begin{equation}
\mathbf{x}^{(i)}(t) = \varphi_{\mathbf{x}_0}(t - t_0) + \bm{\epsilon}_{t}^{(i)} = \varphi(t_i) + \bm{\epsilon}_{t}^{(i)}
\label{eqn:ContDefX}
\end{equation}

Assuming we use $\hat{\mathscr{L}}_{LS}$ as our estimation for $\mathscr{L}$ then we solve:
\begin{displaymath}
\min_{\gamma} \int_{t_0}^{t_f} \sum_{i=1}^n d\left(\gamma(t), \mathbf{x}(t)\right)^2 \norm{\gamma'(t)}\; dt,
\end{displaymath}
subject to $\mathbf{g}$. Here $\norm{\gamma'(t)}$ accounts for the length of $\gamma$ on $M$ so that geodesic trajectories are preferred. At any time instant $t$, the value $\gamma(t)$ minimizing $d\left(\gamma(t), \mathbf{x}^{(i)}(t)\right)^2$ is:
\begin{displaymath}
\gamma^*(t) = \frac{1}{n}\sum_{i=1}^n \mathbf{x}^{(i)}(t) = \varphi(t) + \frac{1}{n}\sum_{i=1}^n \epsilon^{(i)}_t
\end{displaymath}
from \cref{eqn:ContDefX}. (To see this, note the integrand is simply the energy function for a mechanical equilibrium point.) We assumed $\bm{\epsilon}_t^{(i)}$ was unbiased. Therefore, as $n \to \infty$, 
\begin{displaymath}
\frac{1}{n}\sum_{i=1}^n \bm{\epsilon}^{(i)}_t \to \mathbf{0}
\end{displaymath}
and $\gamma^*(t) \to \varphi(t)$. We ignored constraints $\mathbf{g}(\gamma,\dot{\gamma}) \leq \mathbf{0}$ only because we can see that $\varphi$ must satisfy these constraints and therefore, asymptotically so will $\gamma$. A similar argument can be made for $\mathscr{L}_{G}$, but it is not as clean, owing to the additional exponential function in the Gaussian.

Using the above results, we see that the proposed technique for filling in missing information in our discretely sampled noisy signal is (in some sense) an optimal one, assuming a stationary Lagrangian. In the case of non-stationarity, the problem is more difficult, and hence the use of heteroskedastic variances $\sigma_i$ (see \cref{eqn:GaussianWells}) related to the time of the observation.  

The inferred point predictor given in \cref{eqn:pcomp} is simple to implement but does not take constraints into consideration. Supposing we know the true feasible region $\Omega$, we quantify how far outside $\Omega$ a point predictor $\mathbf{p}_i$ could be. 
This can be used to determine whether it is appropriate to go through the effort of computing \cref{eqn:argmaxx} or to simply use \cref{eqn:pcomp}.  

As before, let $\Omega \subseteq M$ be the feasible region for the trajectory $\varphi_{\mathbf{x}_0}(t - t_0)$. Without loss of generality, assume $\Omega$ is a proper subset of $M$, so that the feasible region is non-trivial. Let $Y = M \setminus \Omega \neq \emptyset$ be the infeasible or \textit{forbidden} region. Denote by $\overline{\, \, \cdot \,\,}$ the topological closure of a set and denote by $\partial$ the topological boundary of a set.

We show that feasible regions (and hence forbidden regions) are (partially) inferred as a part of Algorithm \ref{algo:Main}. To do this, we will use the \textit{Hausdorff distance}, defined on the power set $2^M$ of $M$ by 
${\rho: (2^M)^2 \to \mathbb{R}^+}$ by
\begin{displaymath}
\rho(S_1,S_2) = \inf \{d(\mathbf{x},\mathbf{y}) : \mathbf{x}\in S_1,\, \mathbf{y} \in S_2 \}
\end{displaymath}
That is, $\rho$ is the smallest distance between points in $S_1$ and $S_2$. When we write $\rho$ with a set and a single point $\mathbf{x} \in M$, we will abuse notation and understand ${\rho(S_1, \mathbf{x}) \overset{\Delta}{=} \rho(S_1, \{ \mathbf{x}\})}$ so that the singleton $\{\mathbf{x}\} \in 2^M$.

Let $Y$ be a fixed forbidden region with closure and boundary denoted as above. Assume the prediction point $\mathbf{p}_i$ is computed with the unconstrained \cref{eqn:pcomp} and the Epanechnikov kernel $K(\mathbf{x})$ with bandwidth vector $\mathbf{h}$. Recall $d$ is the dimension of $M$, and let $\norm{\cdot}$ be the Euclidean metric on $\R^d$. Then:
\begin{equation}
\mathbf{p}_i \in \Omega \cup \{\mathbf{m} \in M: \rho(\partial \overline{Y}, \mathbf{m} ) \leq \norm{\mathbf{h}}+\max_{1\leq i\leq N}\{{\norm{\bm{\epsilon}_{t_i}}}\}\}\label{eqn:theoerrorbound}
\end{equation}
In other words, the distance from any prediction point to the boundary (of the closure) of the forbidden region is at most the magnitude of the worst-case noise plus the magnitude of the bandwidth $\norm{\mathbf{h}}$. If $\mathbf{p}_i \in \Omega$ this is trivial, so we consider the case when $\mathbf{p}_i \in Y\subseteq M$.


Let $K_{\mathbf{h}}(\mathbf{x})$ be the shifted Epanechnikov product kernel 
\begin{displaymath}
K_{\mathbf{h}} (\langle {x_1, x_2, ..., x_d } \rangle) = K\left(\frac{x_1 - x_1^j}{h_1}, \frac{x_2 - x_2^j}{h_2},..., \frac{x_d - x_d^j}{h_d}\right)
\end{displaymath} centered at $\mathbf{x}^{(j)} = \left\langle {x_1^j, x_2^j, ..., x_d^j} \right\rangle $. Then the support of $K_{\mathbf{h}}(\mathbf{x})$ is the parallelepiped:
\begin{displaymath}
[x_{1}^j - h_1, x^j_{1} + h_1] \times [x_{2}^j - h_2, x_{2}^j + h_2] \times ... \times [x_{d}^j - h_d, x_{d}^j + h_d].
\end{displaymath}
The support of $\hat{f}_i$ (the $i^\text{th}$ estimated distribution) is the union of the supports of the individual kernels centered at $\mathbf{x}_i^{(j)}$ for $1 \leq j \leq |H|$, hence there is some point $\mathbf{x} \in \{\mathbf{x}_i^{(j)}\}_{j=1}^{|H|}$ such that ${d(\mathbf{x}, \mathbf{p}_i) \leq ||\mathbf{h}||}$. For the right discrete time point $t$, $\mathbf{x} = \varphi (t) + \bm{\epsilon}_t$ is perturbed by at most $\max_{1\leq i\leq N}\{{\norm{\bm{\epsilon}_{t_i}}}\}$, then 
\[\mathbf{x} \in \Omega_\epsilon \overset{\Delta}{=}\Omega \cup \{\mathbf{m} \in M: \rho(\partial \overline{Y}, \mathbf{m} ) \leq \max_{1\leq i\leq N}\{{\norm{\bm{\epsilon}_{t_i}}}\}\}\]
 since $ \varphi (t) \in \Omega$. 

To maximize $$\sup_{\mathbf{y} \in (\supp \hat{f}_i) \cap Y} \rho (\Omega_\epsilon, \mathbf{y}),$$ that is, to have conditions which allow $\mathbf{p}_i$ to be as far,
away from the boundary of (the closure) of $Y$, while not being in $\Omega$, we need $\mathbf{x}$ to minimize $\rho(Y \backslash \Omega_\epsilon, \mathbf{x})$.
The closest that ${\mathbf{x}}$ could be to $Y \backslash \Omega_\epsilon$ without being in it is if ${\mathbf{x}} \in (\partial \overline{Y \backslash \Omega_\epsilon} )\cap \Omega_\epsilon$, the boundary of $\overline{Y \backslash \Omega_\epsilon}$. This, of course, assumes that $Y \backslash \Omega_\epsilon$ is open - if it were closed, then $\rho(Y \backslash \Omega_\epsilon,{\mathbf{x}})$ is strictly positive and our argument still holds. We now see that $\rho\left((\partial \overline{Y \backslash \Omega_\epsilon}),\mathbf{p}_i\right) \leq \norm{\mathbf{h}}$, which implies \cref{eqn:theoerrorbound} via the triangle inequality. Assuming that the noise is sufficiently small to allow all observations to be in the feasible region $\Omega$, then $\norm{\mathbf{h}}$ alone serves as an upper bound on the distance inside $Y$ at which $\mathbf{p}_i$ may appear.

It should be noted that the essence of this argument extends to any Kernel whose support is bounded. We also note that if the topological diameter of a forbidden region $Y^\prime$ is smaller than $||\mathbf{h}||$, then this property does not prohibit $\mathbf{p}_i$ from being at any point of $Y^\prime$. 

\section{Experimental Results}\label{sec:Empirical}

We discuss two sets of experiments to test Algorithm \ref{algo:Main}. In the first experiment, we forecast two cruise ships over several days (Carnival line's \Freedom  and \Dream) to evaluate the performance of Algorithm \ref{algo:Main} in a real world context. In the second experiment, we evaluate the efficacy of Algorithm \ref{algo:Main} with three synthetic data sets. Use of a synthetic data set allows us to more closely control the underlying dynamical system and provides a method for exploring potential limitations of the proposed technique.

We use two error metrics to evaluate the algorithm: \textit{absolute pointwise error} ($\mathrm{APE}$), and \textit{percent in highest density region} ($\mathrm{\%HDR}$). Let $\mathbf{x}_i$ be the true position of particle $s$ at time $t_i$. We create a forecast $\mathscr{F}$ using Algorithm \ref{algo:Main} (including information prior to $t_i$), and obtain prediction point $\mathbf{p}_i$ for time $t_i$. The $\mathrm{APE}$ function is defined as the distance $\mathrm{APE}(t_i)\overset{\Delta}{=} d(\mathbf{x}_i, \mathbf{p}_i)$. As noted, we can construct HDR $R_i$ at $t_i$. Let:
\begin{displaymath}
\chi_{R_i}(\mathbf{x}) = \begin{cases} 1 & \mathbf{x} \in R_i \\ 0 & \text{otherwise} \end{cases}
\end{displaymath} 
be an indicator function, then define 
\begin{displaymath}
\mathrm{\%HDR} (\mathscr{F}) \overset{\Delta}{=} \frac{1}{\overline{q}}\sum_{i = N+1}^{N + \overline{q}} \chi_{R_i}(\mathbf{p}_i). 
\end{displaymath}

$\mathrm{APE}$ tells us how far off the pointwise forecast is while $\mathrm{\%HDR}$ tells if the true position is in the derived uncertainty region. We compute mean and standard deviation of $\mathrm{APE}$ for an entire forecast, to give a global error metric for the forecast as a whole.

\subsection{Ship Track Forecasts}


Cruise ships exhibit highly recurrent behavior as they travel from port to port, according to a list of destinations which appeal to tourists. Cruise ships also use AIS to give their positions at a high sampling rate with low noise. This makes them excellent subjects on which to test our algorithm, as we can downsample a portion of a known track and generate noise to create training data. After creating a forecast from this sparse, noisy training data, we can then use the remainder of the track as high-resolution ground truth for an error analysis of the forecast.

In order to test our algorithm with this data, we used (approximately) two-years of positional data on the Carnival line cruise ships \emph{Dream} (from December 2011 - July 2012) and \emph{Freedom} (from March 2012 - June 2013). The data was taken from \cite{sailwx}, under fair use. This data was densely sampled, giving a location for each ship on average about once every fifteen minutes. The first 80\% of the historical trajectory of each ship was used as the historical data for ``training'' the KDE model and the last 20\% was used as an unseen track on which to test. 

We downsampled the training data (the first 80\% for each ship) to give one position every day with exactly 24 hour intervals, while retaining the resolution of the unseen track. Since there were usually not AIS positions at exactly 24 hours of time-difference, we linearly interpolated between the nearest known position before the 24-hour mark and nearest known position after the 24-hour mark. The choice of linear interpolation is ``wrong'' in the sense that we are working on an oblate spheroid as the manifold, but served the purpose of introducing noise into the training data. The result was a sparse noisy track; this was the desired condition for our historical data. 

We generated several forecasts with different parameters. In particular, we considered forecast windows of one week with 15 minute resolution, and with input search radii of 10NM, 20NM, and 40 NM. In each case we chose a bandwidth of 1.5 degrees of latitude/longitude, or 90 nautical miles. The bandwidth was chosen by trial-and-error to yield a smooth forecast. We consider the problem of automated bandwidth selection as a problem for future work.

\begin{figure}
	\subfloat[\emph{Dream} with 10NM search radius]{ \includegraphics[width=0.8\columnwidth]{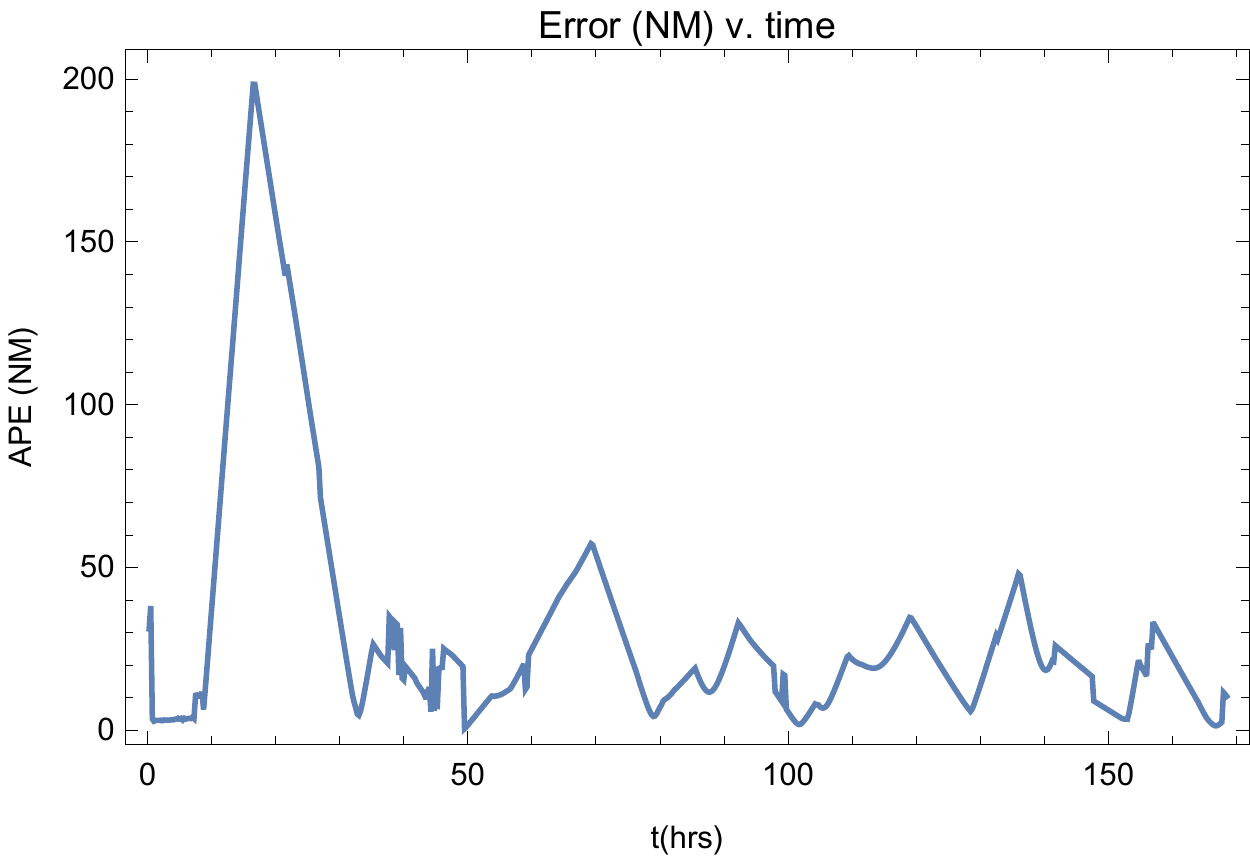}}\\
	\subfloat[\emph{Freedom} with 40NM search radius]{ \includegraphics[width=0.8\columnwidth]{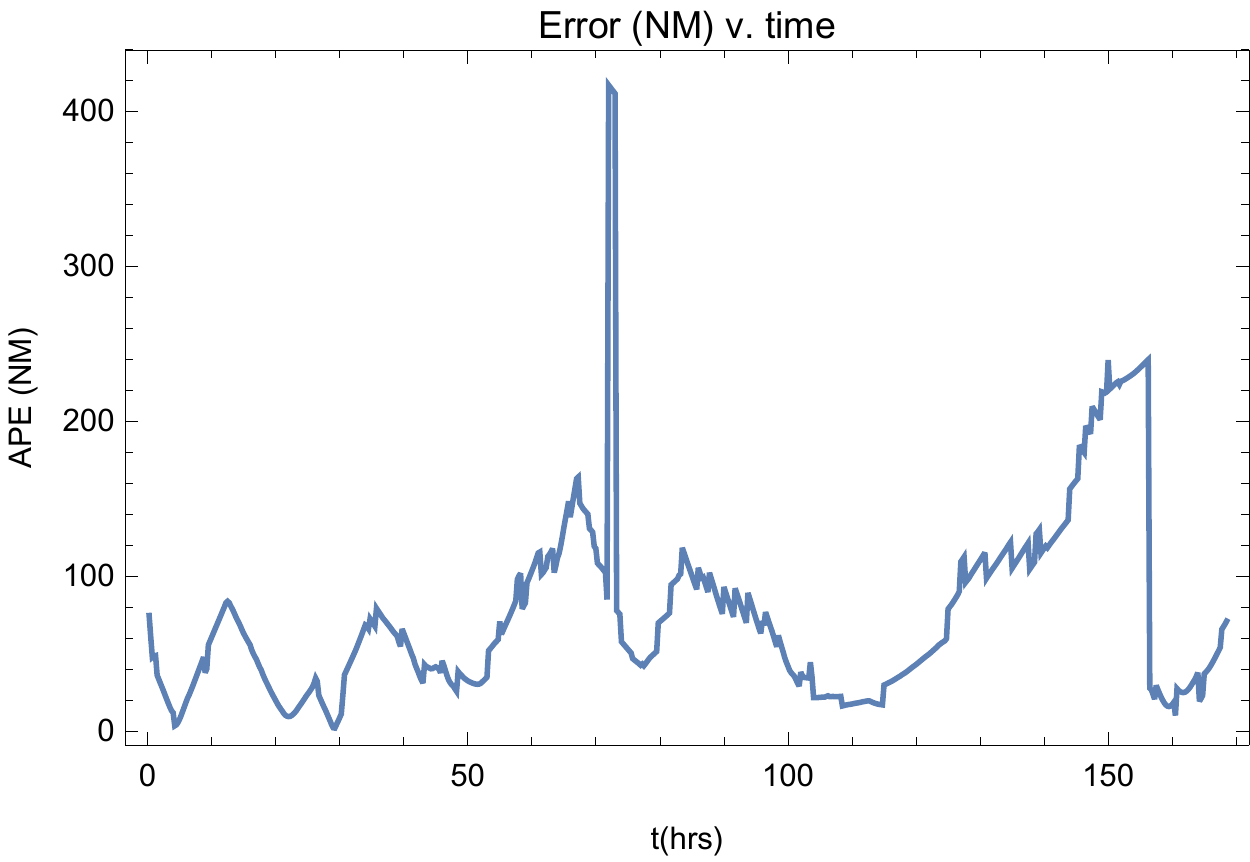}}
\caption{Error plots of pointwise error at 15 minute intervals for 7-day forecast of \emph{Dream} and \emph{Freedom} with search radii of 10NM and 40NM.}\label{fig:SelectedResults}
\end{figure}

After performing the energy minimization step of Algorithm \ref{algo:Main}, we have densely sampled paths. Error metrics for forecasting are are plotted in \cref{fig:SelectedResults}. The trajectories of cruise ships change frequently (e.g., as a result of stop over at ports of call). By way of comparison, we note that course change dynamics would have to be known \textit{a priori} when using (e.g.) a Kalman Filter.

%

\begingroup
\squeezetable
\begin{table*}
\centering
\begin{tabular}{|c|c|c|c|c|}
\hline
Ship & Search Radius & Average Error (NM) & Standard Deviation of Error (NM) & \% in HDR\\
\hline
\emph{Dream} & 10 & 35.9 &43.9 & 90.9\\
\hline
\emph{Dream} & 20 & 65.6 & 125.5 & 89.7 \\
\hline
\emph{Dream} & 40  & 74.8 & 128.5 & 88.7 \\
\hline
\emph{Freedom}& 10  & 93.9 & 66.7 & 71.7 \\
\hline
\emph{Freedom} & 20  & 90.5 & 68.3 & 76.3 \\
\hline
\emph{Freedom} & 40  & 85.9 & 69.9 & 78.9 \\
\hline
\end{tabular}
\caption{Table of results for cruise ship data}\label{table:Results}
\end{table*}
\endgroup

\cref{table:Results} shows summary statistics for the cruise ship forecasting experiments. For \emph{Dream}, as we increase the search radius, we see an increase in the average error and standard deviation of error, and a decrease in percent in HDR. For \emph{Freedom}, the average error went down from 10NM to 20NM to 40NM. The standard deviation also only marginally increased and \%HDR increased significantly. 

It is curious that the examples exhibit different behavior as the search radius grows. One possible explanation for this phenomenon is that for \emph{Dream} with radius 10NM, the prediction in the first several hours is less than 10NM, while for \emph{Freedom} with radius 10NM, there are predictions made in the first hour with error greater than 10NM. By declaring a search radius we are saying two points are in the same location if they are sufficiently close; i.e., we are creating an equivalence class on the observed data. Thus, we cannot expect our error to be smaller than our search radius. If the error \emph{is} initially smaller than the search radius, then expanding the search radius would add extra data, which would contribute to a less accurate prediction. On the other hand, if the error is greater than the search radius, increasing the radius does not add data that is further away than the error, and so it might improve the forecast.

At a higher level, this example provides relevant information about the proposed method. First, there is not necessarily a single best initial search radius - it is context dependent; i.e., a parameter of the model that must be fit. Secondly, it validates our choice of HDR as uncertainty region, because our worst prediction (\emph{Freedom}, 10NM), was still within the HDR $71.7\%$ of the time. Moreover, \cref{table:Results} shows that greater error and standard deviation of error corresponds to a lower $\%HDR$. It is true that the HDR depends on the bandwidth parameter, but with a properly chosen bandwidth, we have a reasonable uncertainty region.

Another positive aspect of our forecast is how it treats land. For the most part, the forecast respects the fact that it must remain in the water, and in the few cases were the forecast does go over land, it is only over small islands or tips of peninsulas, which may not be considered by the energy minimization step \footnote{This step was carried out using a gridded representation of the Earth using a numerical approximation of the line integral.}. This is important to note because we did not give as an input the location of landmasses in the statistical model. Not only does it respect navigational constraints in this manner, but we also see two interesting patterns in the \emph{Freedom} forecast. When the ship goes around the Bahamas, the true track goes west of the islands, while the forecast goes mostly east of the islands. In this sense, we have a valid navigational pattern given by the forecast. A similar occurrence happens as the ship proceeds southeasterly, north of Puerto Rico. The forecast goes south in between Hispaniola and Puerto Rico \emph{avoiding both landmasses}, before proceeding northwest at which point the error goes down to around 50NM. While the forecast was wrong, it gave valid outputs with all knowledge of land contained in the estimate $\hat{\mathscr{L}}$ and upsampling step, and no knowledge of land in computing $\hat{f}_i$ or $\mathbf{p}_i$. This indicates if the data was not sparse (e.g. streaming AIS positions), then the forecast would have avoided land without explicitly computing an approximation $\hat{\mathscr{L}}$.

\begin{figure}
	\includegraphics[width=0.9\columnwidth]{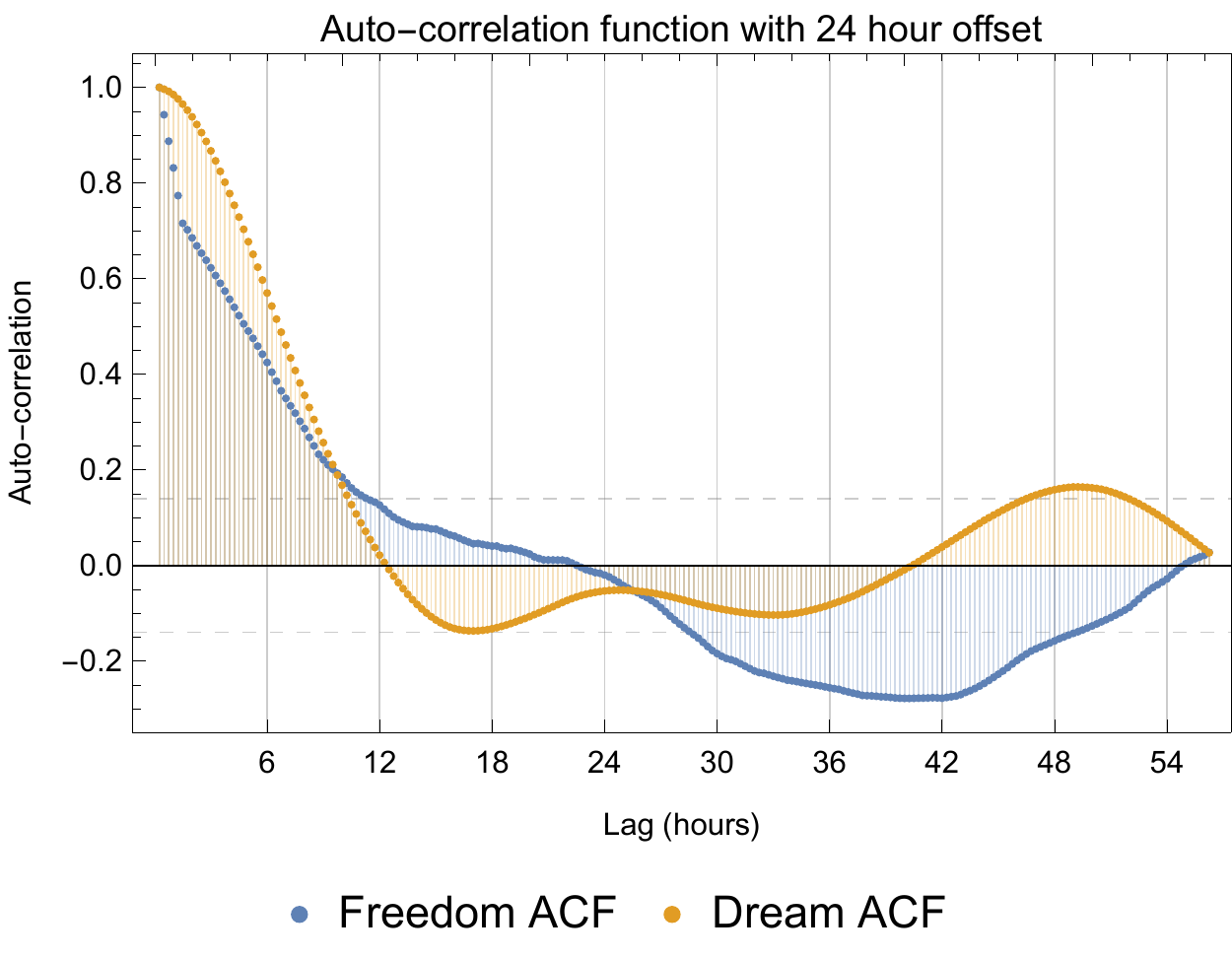}
	\caption{Auto-correlation function of error, with offset of 24 hours. Dashed line indicates critical value for statistical significance at $\pm 0.140$.}\label{fig:acf}
\end{figure}

Examining the plots in \cref{fig:SelectedResults} more closely, we do not see a general trend in error with respect to time. For \emph{Dream} the worst error occurs within the first two days of prediction and is almost entirely temporal. For \emph{Freedom} the worst error is in the last two days and is almost entirely spatial. A traditional forecaster such as a Kalman or particle filter would be expected to have increasing error over time. Finally, we note that the error seems to be periodic. More specifically, it seems to cycle roughly in relation to days. It is possible that this is an artifact of our daily downsampling while preprocessing the historical data. This claim is supported by computing the auto-correlation function for the error time-series, which can be seen in \cref{fig:acf}. The plot shows statistically significant auto-correlation around the 48 hour period for both forecast, as well as for \textit{Dream} at around the 80 hour and 125 hour marks.

To summarize: Algorithm \ref{algo:Main} gives a very reasonable forecast, with only minimal information about the manifold of interest. It respects navigational constraints, and does not appear to loose accuracy over time in any general way.

\subsection{Synthetic Data Forecasts}\label{sec:SynthForecast}


\begin{figure}[t]
\centering
	\subfloat[First synthetic trajectory\label{fig:synthtracks1}]{ \includegraphics[width=0.8\columnwidth]{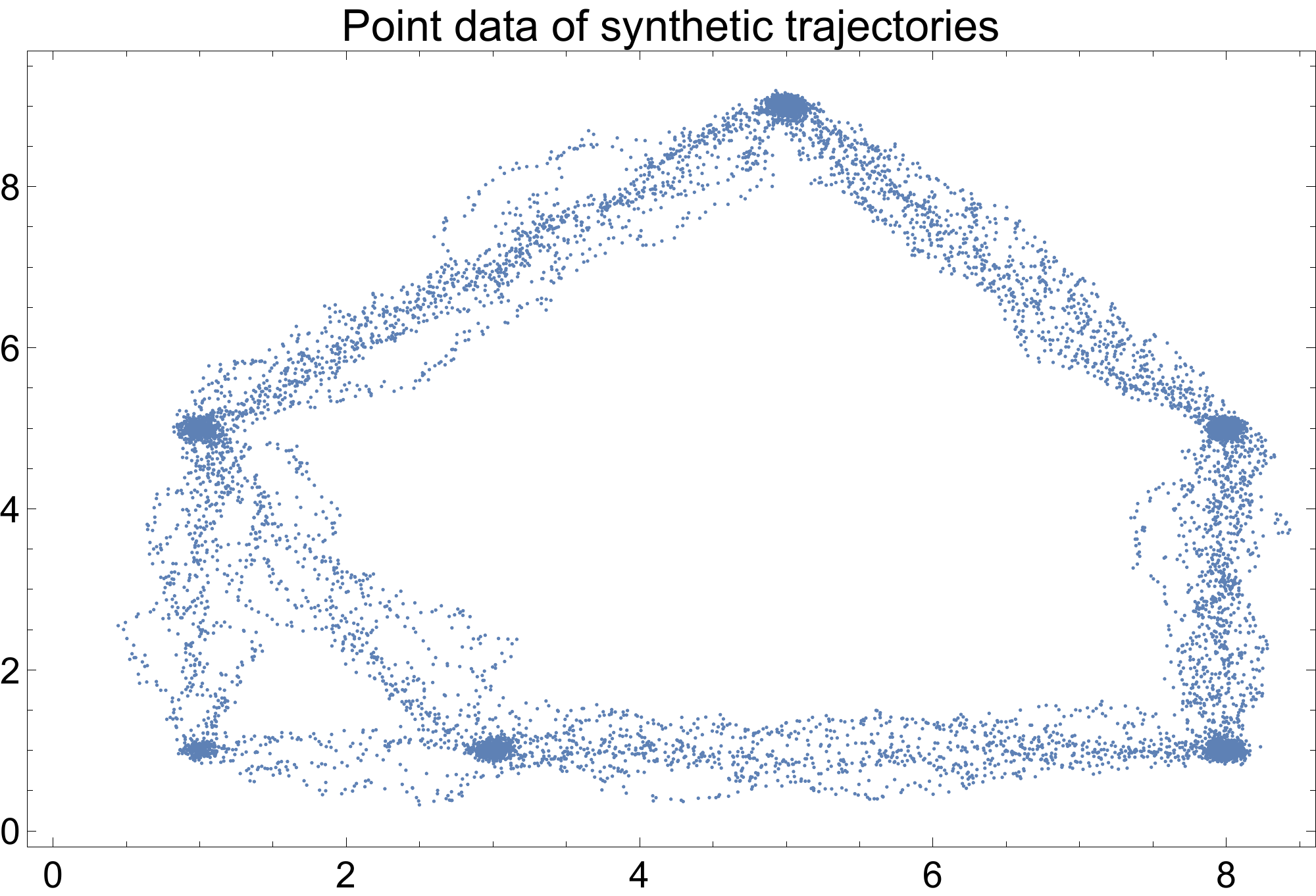}}\\
	\subfloat[Second synthetic trajectory\label{fig:synthtracks2}]{ \includegraphics[width=0.8\columnwidth]{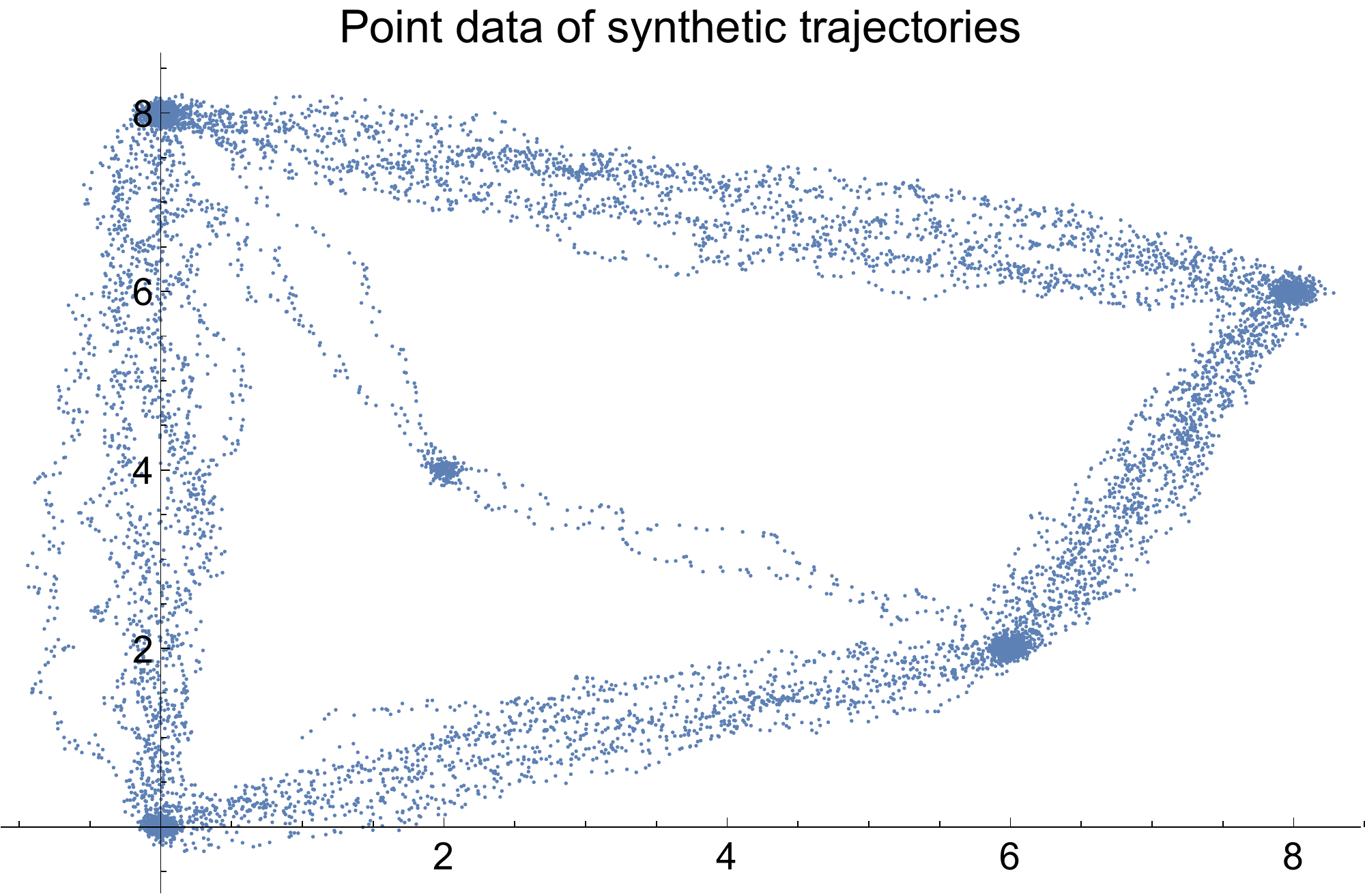}}
\caption{Synthetic trajectory moving between loiter points}
\label{fig:synthtracksData}
\end{figure}

In order to have more control over both the training and testing data set (thereby guaranteeing the data meets our assumptions), we created three synthetic histories of trajectories. The first two were trajectories on $\mathbb{R}^2$, and included loiter points and followed the same underlying model, with a different geometry. The third was the Lorenz system in $\mathbb{R}^3$. This enabled us to compare our results with those in \cite{BGH15}.

\subsubsection{Plane trajectories}\label{sec:Plane}
The first two synthetic histories of trajectories on $\R^2$ consisted of 10,000 data points with Gaussian noise, shown in \cref{fig:synthtracks1} and \cref{fig:synthtracks2}. The generated tracks included six and five loiter points respectively, including a bifurcating trajectory where, after reaching the top leftmost loiter point the particle uniformly chose to go towards the center of the system or proceed due south. The purpose of the loiter points is to understand how the algorithm treats speed implicitly, while the purpose of the bifurcation is to see how the forecasting algorithm deals with such phenomena. 

\begin{figure}[bp]
\centering
	\subfloat[First synthetic trajectory\label{fig:PredVTest1}]{ \includegraphics[width=0.8\columnwidth]{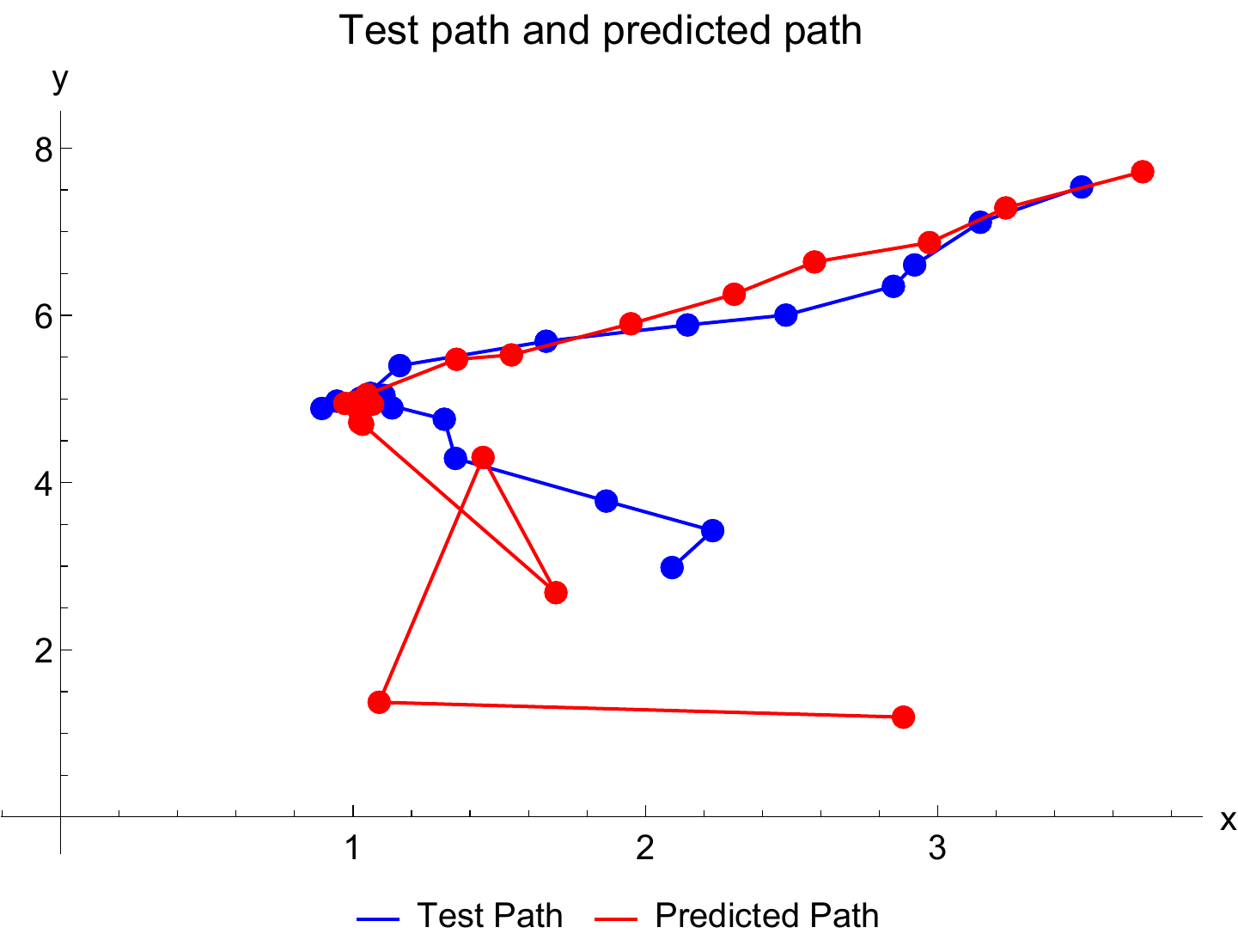} }\\
	\subfloat[Second synthetic trajectory\label{fig:PredVTest2}]{\includegraphics[width=0.8\columnwidth]{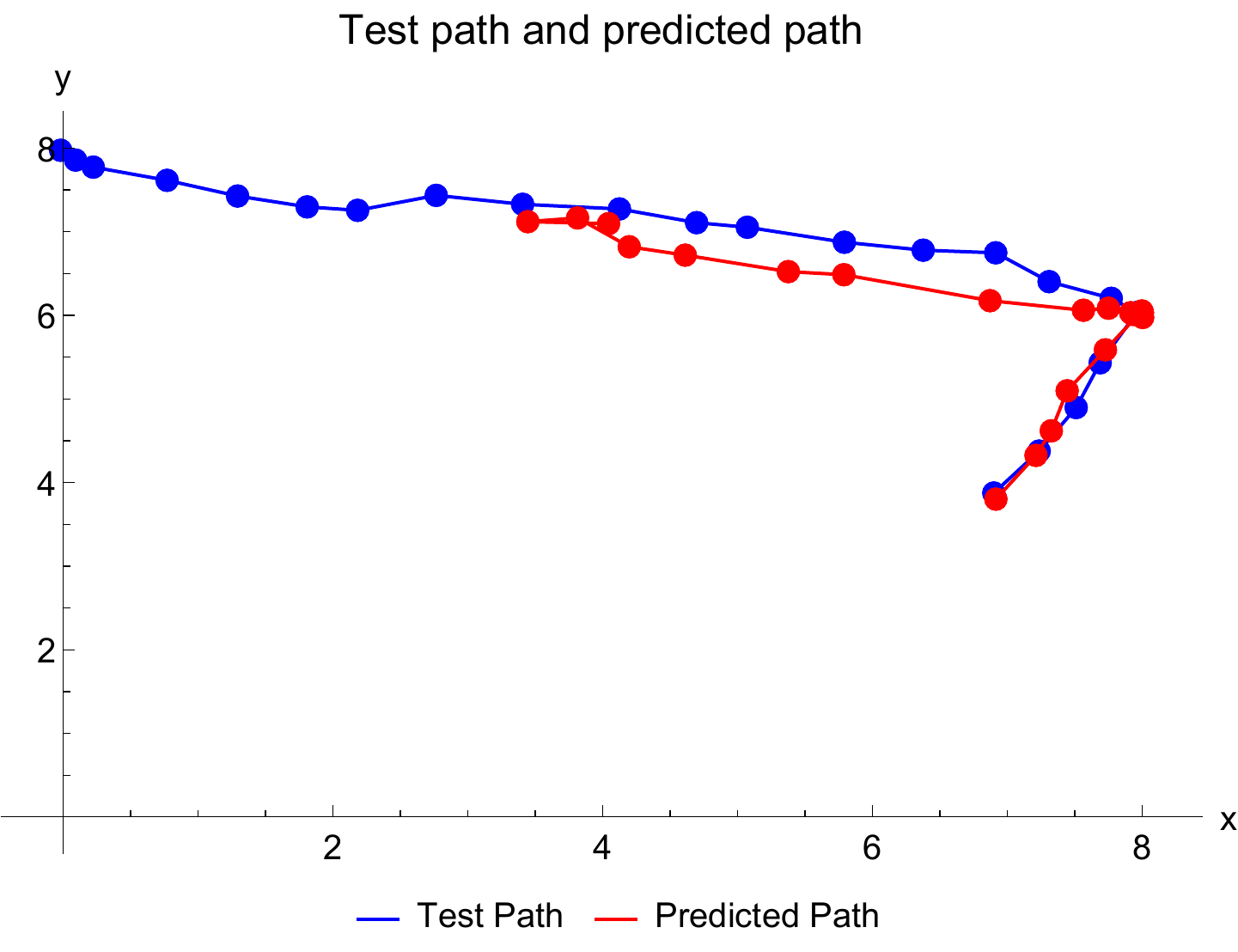} }
\caption{Synthetic trajectories and forecasts moving between loiter points.
}
\label{fig:synthtracks}
\end{figure}

In both cases, we forecasted a length $\overline{q} = 20$ trajectory $\{\mathbf{x}_i\}_{i = N+1}^{N+\overline{q}}$ for the dynamical system that generated the first $N = 10,000$ data points, with $\mathbf{x}_{N+1} = (3.5,7.5)$. In reporting (see \cref{table:SynthError}) we consider each index to be a half-unit of time, so that $\mathbf{x}_{N+1}$ corresponds to reporting time $t = 0.5$ and $\mathbf{x}_{N+\overline{q}}$ corresponds to time $t = 10.5$. We note that $\{\mathbf{x}_i\}_{i = N+1}^{N+\overline{q}}$ was not used to train the model (in the sense of contributing to the data used to build a KDE). These tracks are shown in blue in \cref{fig:PredVTest1} and \cref{fig:PredVTest2}, where the points represent the actual $\{\mathbf{x}_i\}$ and the points are connected linearly in both space and time to give a position for any $t \in [t_{N+1},t_{N+\overline{q}}] \subseteq \mathbb{R}^+$.

After generating training data and a ground truth trajectory for comparison to predictions, we compute a prediction $\{\mathbf{p}_i\}_{i=N+1}^{N+ \overline{q}}$. We plot the predictions in red in \cref{fig:PredVTest1} and \cref{fig:PredVTest2}.

\begin{table}[htbp]
\footnotesize
\begin{tabular}{|l|| c | c || c | c |}
\hline
\text{Time} & \begin{tabular}{c} Prediction Error\\ 1st Test \end{tabular} &\begin{tabular}{c} In 70\%\\ HDR?\end{tabular} & \begin{tabular}{c} Prediction Error\\ 2nd Test \end{tabular} & \begin{tabular}{c} In 95\%\\ HDR?\end{tabular}\\
\hline
 0.5 & 0.277079 & \text{Yes} & 0.0748012 & \text{Yes} \\
\hline
 1.0 & 0.192594 & \text{Yes} & 0.056595 & \text{Yes} \\
\hline
 1.5 & 0.271879 & \text{Yes} & 0.333663 & \text{Yes} \\
\hline
 2.0 & 0.39765 & \text{Yes} & 0.415056 & \text{Yes} \\
\hline
 2.5 & 0.305548 & \text{Yes} & 0.50537 & \text{No} \\
\hline
 3.0 & 0.194307 & \text{Yes} & 0.272652 & \text{Yes} \\
\hline
 3.5 & 0.202291 & \text{Yes} & 0.786428 & \text{No} \\
\hline
 4.0 & 0.207912 & \text{Yes} & 1.33602 & \text{No} \\
\hline
 4.5 & 0.122324 & \text{Yes} & 1.73251 & \text{No} \\
\hline
 5.0 & 0.0112425 & \text{Yes} & 2.33992 & \text{No} \\
\hline
 5.5 & 0.0681667 & \text{Yes} & 3.02051 & \text{No} \\
\hline
 6.0 & 0.0993839 & \text{Yes} & 3.21997 & \text{No} \\
\hline
 6.5 & 0.128976 & \text{Yes} & 3.64753 & \text{No} \\
\hline
 7.0 & 0.139858 & \text{Yes} & 3.65211 & \text{No} \\
\hline
 7.5 & 0.333929 & \text{Yes} & 3.16792 & \text{No} \\
\hline
 8.0 & 0.107408 & \text{Yes} & 3.27594 & \text{No} \\
\hline 
 8.5 & 0.274433 & \text{Yes} & 2.86188 & \text{Yes} \\
\hline
 9.0 & 2.10894 & \text{Yes} & 2.9664 & \text{Yes} \\
\hline
 9.5 & 0.0949259 & \text{Yes} & 3.07647 & \text{Yes} \\
\hline
 10.0 & 2.52883 & \text{Yes} & 3.28763 & \text{No} \\
\hline
 10.5 & 2.32441 & \text{Yes} & 4.15615 & \text{No} \\
 \hline
\end{tabular}
\normalsize
\caption{$APE$ results for forecasts of first and second trajectories}\label{table:SynthError}
\end{table}

In \cref{fig:PredVTest1}, we see that the overall shape of the prediction and ground truth are roughly similar, and they are extremely similar in \cref{fig:PredVTest2}. In particular, the predicted path and ground truth are very close up through the predictions made after the ground truth leaves the loiter regions at $(1,5)$, and $(8,6)$ respectively. 

For the first forecast,
the average $\mathrm{APE}$ is $0.495$ with a standard deviation of $0.773$ units. It is clear that this relatively large deviation comes from only $t=9,10,10.5$. Looking only from $t=0.5$ to $t=8.5$, we see the $\mathrm{APE}$ is less than $0.4$ units, and is as low as $0.0112$ units at $t= 5$. Moreover, every observation is within a 70\% HDR. 

For the second forecast, the average $\mathrm{APE}$ is $2.104$ with a standard deviation of $1.414$ units. Interestingly while the error is not monotonically increasing, the error is below average for every point in the first half of the trial, and above average for the second half of the trial. We also see drastically different performance with regards to the HDR. Only $8$ of the $20$ points lie within a 95\% HDR. We chose to report the 95\% HDR to test the importance of the choice of $\alpha$.
The choice of 95\% is also in keeping with the standard 95\% confidence interval for a test-statistic.

The first notable failure in the proposed method can be observed in the first trajectory between times $t = 8.5$ and $t = 9$. The predictions linger at the loiter point for a slightly longer time than ground truth. After there is a prediction made outside of the loiter region ($t=9$), the distance to $\mathbf{p}_{N+17}$ from $\mathbf{p}_{N+16}$ is much larger than the distance between any two points on the test path. We see this phenomenon to a smaller degree between $\mathbf{p}_{N+13}$ and $\mathbf{p}_{N+14}$ in the second prediction.

To some extent this phenomenon is negative, because in some cases (as in the first trajectory) this may result in a prediction that is traveling too fast; i.e., the resulting predicted trajectory may not be physically realizable.
On the other hand, the observed behavior is positive because the algorithm self-corrects, in the sense that after a loiter point, it catches up to where it should be, rather than just proceeding on from the loiter point at a constant velocity. If it did proceed from the loiter point at a constant velocity, it would then be temporally behind the correct position regardless of how accurate the prediction was spatially. 

We do note that in the second trajectory, error is almost entirely temporal. For both cases the spatial nearness threshold set for the algorithm was $\epsilon=0.25$. In \cref{table:SynthError2b} we have computed the distance between every predicted point $\mathbf{p}_i$ and the closest point on the true trajectory. This distance ignores the time at which a forecast expects the particle to be in a certain location. 

In \cref{table:SynthError2b}, we see that mean distance between a forecast point and any true point is less than $\epsilon$, and 61.9\% of the individual points were less than $\epsilon$ away from the nearest point on the forecast. We draw attention to this fact because by choosing $\epsilon$, as we noted, we are essentially constructing an equivalence relation on the data. 
Thus, the majority of the forecast points were in the ``same place'' as at least one of the true points.
\begin{table}[htbp]
\begin{tabular}{|l||c|c|}
\hline
Time &\text{Distance} & \begin{tabular}{c}Less than\\$\epsilon =.25$\end{tabular}\\
\hline
 0.5 & 0.0748012 & \text{Yes} \\
 \hline
 1.0 & 0.056595 & \text{Yes} \\
 \hline
 1.5 & 0.25832 & \text{No} \\
 \hline
 2.0 & 0.211505 & \text{Yes} \\
 \hline
 2.5 & 0.159474 & \text{Yes} \\
 \hline
 3.0 & 0.0399122 & \text{Yes} \\
 \hline
 3.5 & 0.0406217 & \text{Yes} \\
 \hline
 4.0 & 0.0733248 & \text{Yes} \\
 \hline
 4.5 & 0.0401314 & \text{Yes} \\
 \hline
 5.0 & 0.0283953 & \text{Yes} \\
 \hline
 5.5 & 0.0472536 & \text{Yes} \\
 \hline
 6.0 & 0.119545 & \text{Yes} \\
 \hline
 6.5 & 0.250742 & \text{No} \\
 \hline
 7.0 & 0.493783 & \text{No} \\
 \hline
 7.5 & 0.388359 & \text{No} \\
 \hline
 8.0 & 0.544281 & \text{No} \\
 \hline
 8.5 & 0.397532 & \text{No} \\
 \hline
 9.0 & 0.458194 & \text{No} \\
 \hline
 9.5 & 0.328007 & \text{No} \\
 \hline
 10.0 & 0.211558 & \text{Yes} \\
 \hline
 10.5 & 0.193224 & \text{Yes} \\
 \hline
\end{tabular}
\caption{Error between predicted point at time $t$ and the nearest true point ignoring time for the trial in \cref{fig:PredVTest2}}\label{table:SynthError2b}
\end{table}

The second notable failure occurs $t = 9.5$ in the first trajectory and $t = 10.5$ in the second trajectory, where the prediction back-tracks toward the loiter point, rather than remaining on its present trajectory. This is not consistent with any behavior exhibited by the ground-truth particle at any point on the trajectory. As with the first failure, there are both positive and negative aspects to this phenomenon. On the one hand, for the first trajectory the $\mathrm{APE} = 0.095$ at $t=9.5$ is quite low. On the other hand, in both trajectories, the back-tracking movement is not physically representative of any pattern occurring in the data.

\begin{figure}[htbp]
\centering
	\subfloat[First synthetic trajectory\label{fig:predcontour1}]{ \includegraphics[width=0.85\columnwidth]{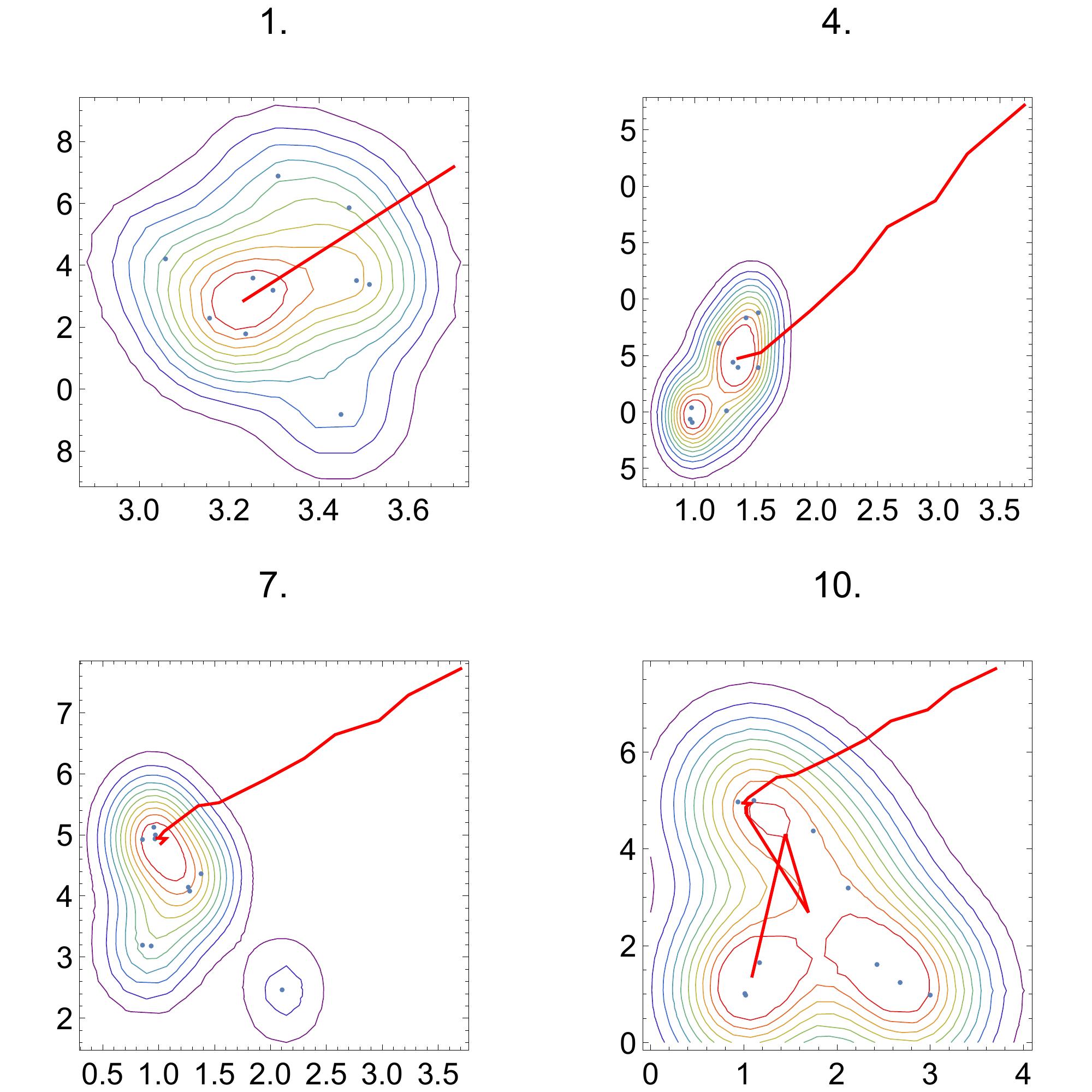} }\\
	\subfloat[Second synthetic trajectory\label{fig:predcontour2}]{\includegraphics[width=0.85\columnwidth]{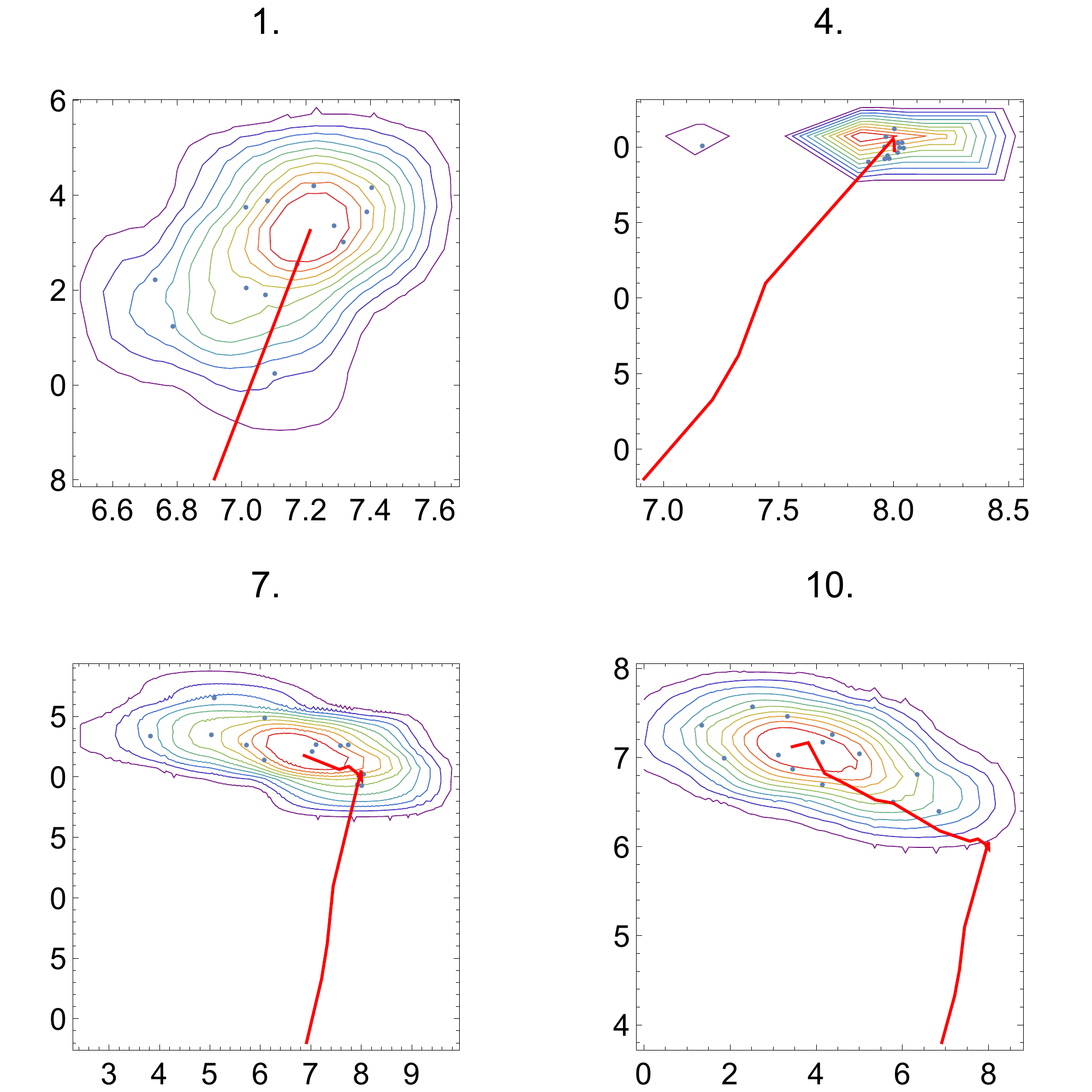} }
\caption{Contour plot of KDE $\hat{f}_i$ at time $t_i$ with predicted track and historical data points overlaid.}
\label{fig:synthtracksContour}
\end{figure}

We note that the first predicted path correctly used the right fork of the bifurcating path. This is coincidental as the underlying model chose uniformly to move to the right or down. The prediction points happened to follow the correct path because there was a slight skew in the data toward the correct fork and then the test path also went right. We note that the underlying data used to train the KDE contained information on the bifurcation and the uncertainty included both possible paths. In the proposed algorithm, the uncertainty regions are far more important than specific pointwise predictors. This can be seen concretely in the multi-modal distribution of \cref{fig:predcontour1} for $t=10$.

The third notable failure also occurs at time $t = 10$ for the first trajectory. We see the prediction is made at the loiter point $(1,1)$ corresponding to the left fork. This makes sense because the distance between the loiter point at $(1,5)$ and $(1,1)$ is less than the distance between $(1,5)$ and $(3,1)$. In other words, at the time when a particle following the left fork would be at the loiter point, a particle following the right fork would along the line between $(1,5)$ and $(3,1)$ (approximately). The concentration of mass of historical points (corresponding to the greatest peak in the KDE) at this time should then be expected to occur, as is the case, at $(1,1)$, because the historical points on the right fork are more spread out. We can see this shown in the contour plot of the KDE at time $10$ in \cref{fig:predcontour1}. This phenomenon also occurs (switching the role of the two forks) at time $t = 10.5$, where we see the prediction is made near $(3,1)$, another loiter point. This phenomenon does not occur in the second trajectory because the region in which the trajectory was forecat does not include any bifurcating portion of the track.

We note, it is relatively straightforward to construct a multi-hypothesis pointwise forecast \cite{GBS16} that would provide branching paths following multiple local maxima of the KDE. This is considered in future work.


With the exception of the three errors noted, the predictions made were quantitatively and qualitatively good. The prediction was close in space (if not always in time) to the ground truth prior to reaching loiter points and along bifurcations in the path.  Considering that we never explicitly compute speed, this is good performance. We also see that near the loiter points, the error becomes quite low. This makes sense, because we are most sure of the position of the particle when it is at a loiter point. 
Using the membership in the HDR as a measure of performance, we see that it is a good way to measure the temporal accuracy of a forecast, but not necessarily the spatial accuracy. This fact is exemplified by trajectory two. We also note that the HDR does not represent a ``confidence region'' in the frequentist sense of the term, but instead has a more Bayesian flavor.

\subsubsection{Lorenz System}
We use the classic Lorenz system defined by the system of differential equations in $\mathbb{R}^3$: 
\begin{align*}
\frac{d}{dx} &= \sigma (y-x)\\
\frac{d}{dy} &= \rho x - y - xz\\
\frac{d}{dz} &= xy - \beta z. \\
\end{align*}
The Lorenz system is the standard example of a chaotic dynamical system \cite{hirsch-smale-devaney} with a fractal strange attractor. Work on forecasting a particle whose underlying dynamics are governed by the Lorenz system has been considered in \cite{BGH15}. Our goal is this section is to create a forecast using \cref{algo:Main} that can be compared to the work in \cite{BGH15}. We used a Mathematica code snippet \cite{wiki:Lorenz_system} to generate 30000 points from a Lorenz system starting at the point $\langle 1,1,1 \rangle \in \mathbb{R}^3$. The parameters were fixed to those in Lorenz's original paper, with $\sigma = 10,\, \rho = 28,\, \beta = \frac{8}{3}$. 

To the sample of 30000 data points, we added unbiased Gaussian noise. We set the spatial tolerance to $\epsilon = 3$ and for the directional tolerance required the dot product to be positive. This is equivalent to partitioning the space with the plane whose normal vector is the velocity vector at $t_0$. Given these conditions we attempted to forecast the movement of the dynamics for the middle $\frac{1}{3}$ of the original data (prior to adding noise) given the noisy data. Since the Lorenz system is chaotic but deterministic, this is equivalent to to generating a new forecast. We used the na\"{i}ve interpretation of manifold and simply used the embedding in $\mathbb{R}^3$. We note that there may be a better coordinate system to choose. However, finding an ideal manifold in which to embed the strange attractor may be difficult due to its fractal nature. 

\begin{figure}[htbp]
\centering
\includegraphics[width = 0.8\columnwidth]{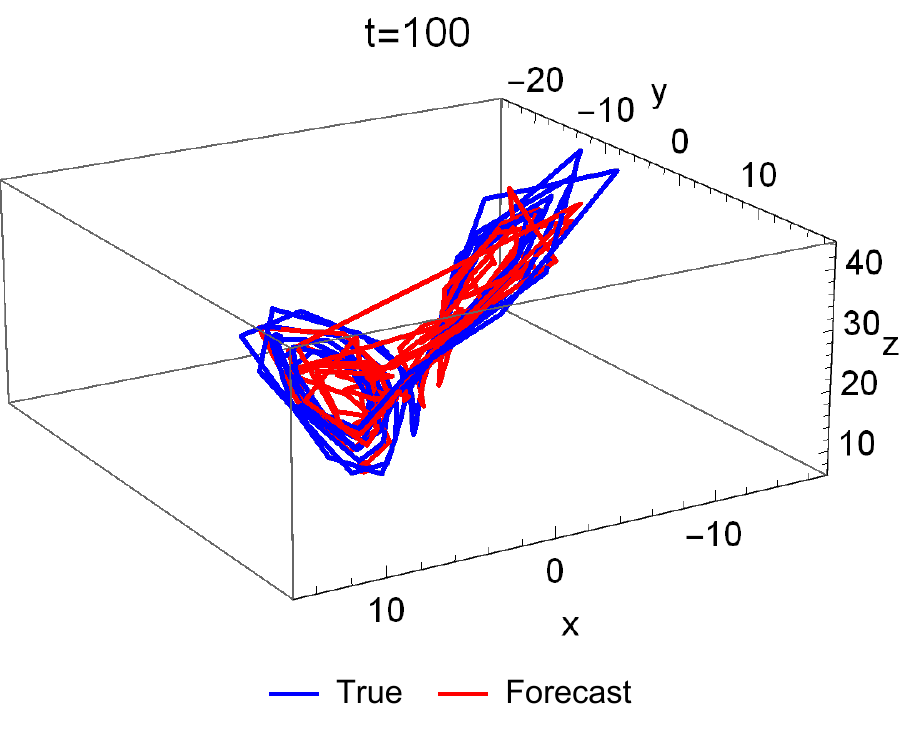}
\caption{Forecast (red) and true path (blue) of a Lorenz system}\label{fig:lorenzforecast}
\end{figure}

\begin{figure}[htbp]
\centering
\includegraphics[width = 0.8\columnwidth]{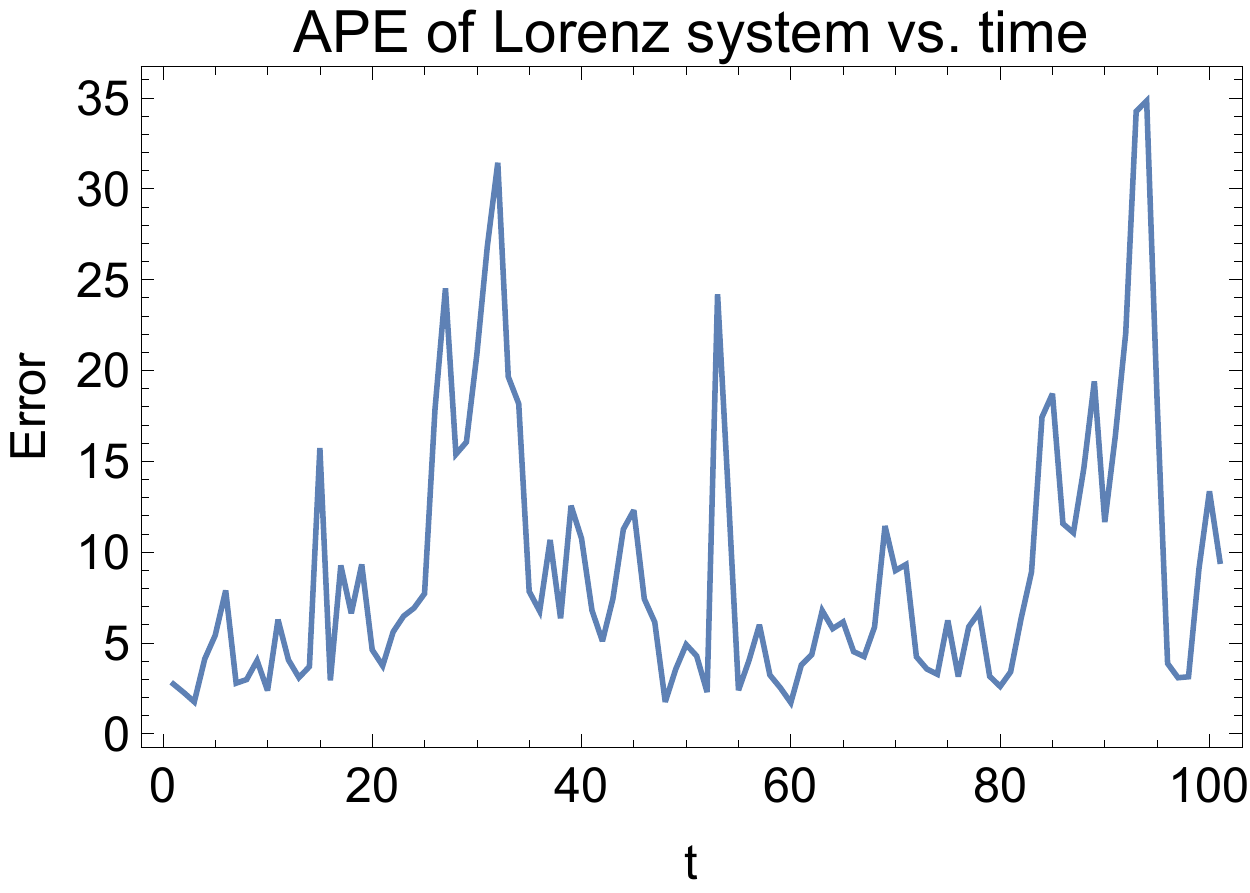}
\caption{Plot of $\mathrm{APE}$ vs. time for a KDE forecast of a Lorenz system.}\label{fig:lorenzerror}
\end{figure}

A visualization of the true path (blue) and our forecast (red) can be seen in \cref{fig:lorenzforecast}.
In \cref{table:lorenzerror} we report the error of every fifth time step for our forecast of the Lorenz system. In \cref{fig:lorenzerror} we show a plot of the full results for all 101 points in time. The mean $\mathrm{APE}$  of the forecast was 9.01 units, with a standard deviation of 7.27. Moreover, 92 out of 101 data points were in the 70\% HDR, even though \cref{table:lorenzerror} has no points outside of the 70\% HDR. 

It is difficult to see in the static image, but by animating the forecast reveals that this forecast has errors which are consistent with the planar forecast in \cref{sec:Plane}, namely that (i) error is often temporal in nature, (ii) the forecast regresses towards the system's center of mass, and (iii) the forecast spends some amount of time on the wrong ``lobe'' of the Lorenz system. We also do not see any strong trend in the error as a function of time. At first glance, the error appears somewhat periodic. This is a reasonable suggestion, due to the pseudo-periodic nature of the movement of a single lobe of the orbit of the Lorenz system. However a discrete Fourier analysis does not support this observation; the error is mostly likely chaotic consistent with the motion of the underlying dynamical system. Moreover, $12.8\%$ of the forecast points have $\mathrm{APE} \leq 3$. This is notable, because we used a spatial tolerance of $3$. As we have previously observed, we cannot reasonably ask that the forecast has error less than that. We also note that the performance of the algorithm with respect to the highest density region computation is quite good. 

\begin{figure}[htbp]
\centering
\includegraphics[width = 0.9\columnwidth]{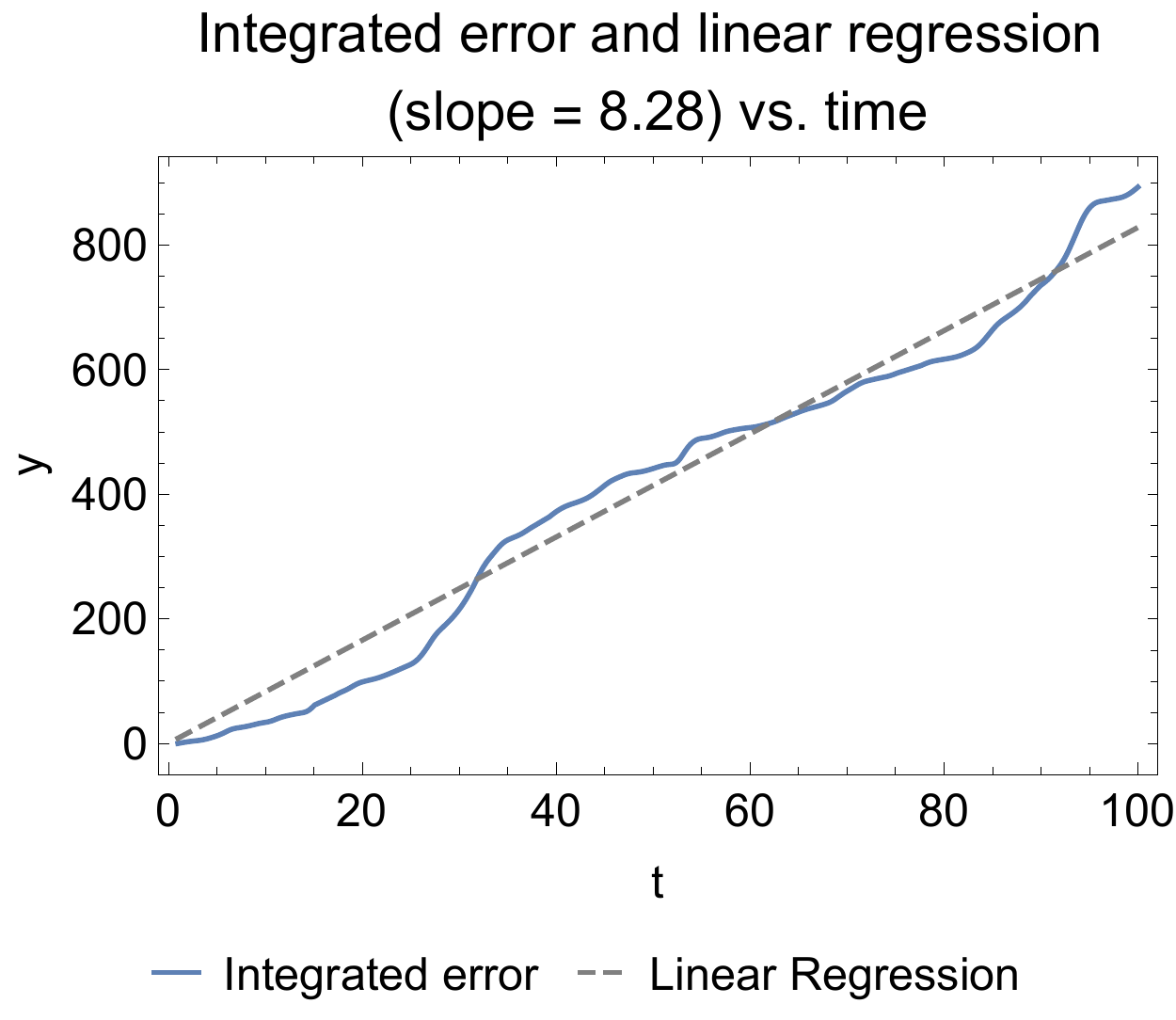}
\caption{Plot of the error curve in \cref{fig:lorenzerror} integrated as a function of time, compared to the line of best fit through the origin.}\label{fig:lorenzerrorintegrated}
\end{figure}

Finally, we make one surprising observation. Consider the error curve shown in \cref{fig:lorenzerror} as a function of time, which we will write $\mathrm{APE}(t)$. We integrate the curve with respect to time; i.e., we numerically compute:
\[ g(t) = \int_0^t \mathrm{APE}(\tau)\; d\tau.\]
The plot of $g(t)$ is shown in \cref{fig:lorenzerrorintegrated}, along with the line of best fit when the $y$-intercept is forced to zero. We see that a line $g_\ell(t)$ is a plausible model for $g$, and moreover that the approximate slope of $g_\ell$ is 8.28. Using all default settings from Mathematica to compute the fit, we have a 95\% confidence interval on the slope of $g_\ell$ of $(8.131,8.425)$. This appears similar in value to the invariant measure computed for the Lorenz system in \cite{BGH15}. Since the invariant measure is the asymptotic error, were it true that a theoretical value of $\dot{g}_\ell$ were the same as the invariant measure $\mathrm{IM}$, then it would be that
\[ \lim_{t \rightarrow \infty} \int_0^t \mathrm{IM} \; d\theta = \lim_{t \rightarrow \infty} g(t). \] 
However, we only conjecture this relationship through experimental evidence rather than any theoretical result.

\begin{table}
\begin{tabular}{|r||l|l|}
\hline
\text{Time} &\begin{tabular}{c} Prediction \\Error\end{tabular} &\begin{tabular}{c} In 70\%\\ HDR?\end{tabular}\\
 \hline
 5 & 5.419 & \text{Yes} \\
 \hline
 10 & 2.376 & \text{Yes} \\
 \hline
 15 & 15.731 & \text{Yes} \\
 \hline
 20 & 4.621 & \text{Yes} \\
 \hline
 25 & 7.712 & \text{Yes} \\
 \hline
 30 & 20.881 & \text{Yes} \\
 \hline
 35 & 7.834 & \text{Yes} \\
 \hline
 40 & 10.752 & \text{Yes} \\
 \hline
 45 & 12.306 & \text{Yes} \\
 \hline
 50 & 4.908 & \text{Yes} \\
 \hline
 55 & 2.394 & \text{Yes} \\
 \hline
 60 & 1.709 & \text{Yes} \\
 \hline
 65 & 6.154 & \text{Yes} \\
 \hline
 70 & 8.984 & \text{Yes} \\
 \hline
 75 & 6.244 & \text{Yes} \\
 \hline
 80 & 2.603 & \text{Yes} \\
 \hline
 85 & 18.719 & \text{Yes} \\
 \hline
 90 & 11.649 & \text{Yes} \\
 \hline
 95 & 18.328 & \text{Yes} \\
 \hline
 100 & 13.352 & \text{Yes} \\
 \hline
\end{tabular}
\caption{Selected $\mathrm{APE}$  for \\ Lorenz system forecast}\label{table:lorenzerror}
\end{table}

\section{Conclusions and Future Directions}\label{sec:Concl}


The previous literature on modeling and forecasting using Kernel Density Estimates has focused on well sampled data. In this paper, we develop Algorithm \ref{algo:Main} to generate forecasts for the cases when we have a sparse, noisy data drawn from a recurrent trajectory. This approach offers an alternative to existing forecasting methods in cases when data are sparse and noisy, since it requires significantly less data. The energy minimization technique we use allows us to ``connect the dots'' between existing data points in an intelligent way, so that we can sample data as finely as we wish.

Algorithm \ref{algo:Main} has several useful features. Traditional forecasters use some approximation or knowledge of speed, which is then projected forward in time. Our forecasting algorithm treats speed implicitly as we develop time-indexed probability distributions over a smooth Riemannian manifold. This simplifies the computation, and makes the algorithm path-independent. In practice, this allows us to pick specific times $t_i$ at which to provide a forecast position $\mathbf{p}_i$ and uncertainty function $f_i$, rather than having the requirement of computing intermediate steps to predict the whole path.

The algorithm also has nice theoretical properties. In particular, the optimal solution of a minimal energy trajectory is the asymptotic limit of the prediction as the sampling rate goes to infinity. Moreover, with a reasonable choice of parameters, it respects the existence of forbidden regions, albeit with a ``fuzzy'' boundary. With reasonable assumptions on (or knowledge of) the noise, we can easily quantify the fuzziness of the boundary.

When we assume but cannot guarantee that the governing energy function is the same, (as in the cruise ship forecasts) we still get reasonably good results. When we \emph{can} guarantee that the underlying energy function is the same (as in our synthetic forecast) the algorithm performs quite well. It gives forecasts, with error that is mostly temporal, rather than spatial. When the error is spatial, it is as a result of a valid path which was not taken. In both the spatial and temporal error cases we still have an uncertainty region for each point, given by the highest density region of the KDE $\fhat$, and experimental evidence supports our claim that this makes sense as an uncertainty region.

While the algorithm performs well, there are certainly improvements which can be made. In particular, we note that often times the error is temporal rather than spatial. That is, the predicted point is not off because the predicted trajectory has high error in space, but rather because the time is off. A next step would be to consider how to improve this weakness in the algorithm. Additionally, for our ship forecast, we heuristically tuned the bandwidth to give a smooth path. It would be desirable to find a non-heuristic way to find the appropriate bandwidth, as the existing bandwidth selection rules (particularly Scott's and Silverman's Rules of Thumb) give a track with high error. Finally, improving our handling of bifurcating tracks when constructing $\mathbf{p}_i$ may also improve the resulting forecasts. 
%

\section{Acknowledgements} This work was supported in part by the Office of Naval Research through Naval Sea Systems Command DO 0451 (Task 23875). We would like to thank Brady Bickel, Eric Rothoff, and Douglas Mercer for helpful conversations during the development of this paper.

\bibliographystyle{apsrev4-1}
\bibliography{KDEForecast}
\end{document}